\documentclass{article} 
\usepackage{iclr2027_conference,times}

\usepackage{amsmath,amsfonts,bm}

\def\Secref#1{Section~\ref{#1}}

\def\eqref#1{equation~\ref{#1}}

\def\1{\bm{1}}

\DeclareMathAlphabet{\mathsfit}{\encodingdefault}{\sfdefault}{m}{sl}
\SetMathAlphabet{\mathsfit}{bold}{\encodingdefault}{\sfdefault}{bx}{n}

\usepackage{hyperref}
\usepackage{url}
\usepackage{graphicx}
\usepackage{booktabs}
\usepackage{multirow}
\usepackage{amsmath,amssymb}
\usepackage{algorithm}
\usepackage{algpseudocode}
\usepackage{subcaption}
\usepackage{wrapfig2}
\usepackage{needspace}
\usepackage[table]{xcolor}
\usepackage{xspace}
\usepackage{longtable}
\usepackage{array}
\usepackage{tabularx}
\usepackage{fvextra}
\DeclareUnicodeCharacter{2605}{\ensuremath{\star}}
\DeclareUnicodeCharacter{2264}{\ensuremath{\le}}
\DeclareUnicodeCharacter{26A0}{[!]}
\DeclareUnicodeCharacter{25CF}{\ensuremath{\bullet}}
\DeclareUnicodeCharacter{25CB}{\ensuremath{\circ}}
\DeclareUnicodeCharacter{2192}{\ensuremath{\rightarrow}}
\DeclareUnicodeCharacter{2194}{\ensuremath{\leftrightarrow}}

\definecolor{groupbg}{gray}{0.92}          
\definecolor{oursbg}{RGB}{232,240,254}     
\definecolor{deltagood}{RGB}{0,128,0}       
\definecolor{deltabad}{RGB}{190,0,0}        
\newcommand{\dgain}[1]{\,{\scriptsize\textcolor{deltagood}{#1}}}  
\newcommand{\dloss}[1]{\,{\scriptsize\textcolor{deltabad}{#1}}}   
\newcommand{\dneu}[1]{\,{\scriptsize\textcolor{gray}{#1}}}        

\newcommand{\compacttables}{%
  \renewcommand{\arraystretch}{0.98}%
  \setlength{\aboverulesep}{0.3ex}%
  \setlength{\belowrulesep}{0.45ex}%
}
\AddToHook{env/table/begin}{\compacttables}
\AddToHook{env/table*/begin}{\compacttables}
\AddToHook{env/wraptable/begin}{\compacttables}
\AddToHook{env/longtable/begin}{\compacttables}

\makeatletter
\renewcommand{\paragraph}{\@startsection{paragraph}{4}{\z@}%
  {0.9ex plus 0.2ex minus 0.1ex}{-1em}{\normalsize\bfseries}}
\makeatother

\newcommand{\modeS}{\mathcal{S}}   
\newcommand{\modeO}{\mathcal{O}}   
\newcommand{\framework}{\textsc{HyWe}\xspace} 
\newcommand{\datapipe}{\textsc{HyWeCollector}\xspace} 
\newcommand{\benchmark}{\textsc{HyWeBench}\xspace} 
\newcommand{\agent}{\textsc{HyWe-35B}\xspace}  
\newcommand{\task}[1]{``\textit{#1}''}                

\title{HyWe: Unifying Search and Operation\\for Real-World Hybrid Web Tasks}

\author{%
\begin{minipage}[t]{\dimexpr\textwidth-2\tabcolsep\relax}
\raggedright\normalfont
{\bfseries
Zijian Cai$^{1,2,3,*}$, Haowei Liu$^{4,*}$, Kailai Chen$^5$,
Lin Peng$^6$, Cong Wan$^6$,\\[2pt]
Ziqi Zhang$^{1,2}$, Chunfeng Yuan$^{1,2,3,\dagger}$, Bing Li$^{1,2,3,7}$,
Weiming Hu$^{1,2,3,8}$, Haiyang Xu$^4$}\\[4pt]
{\small
$^1$Beijing Key Laboratory of Super Intelligent Security of Multi-Modal Information, CASIA\\[1pt]
$^2$State Key Laboratory of Multimodal Artificial Intelligence Systems, CASIA\\[1pt]
$^3$School of Artificial Intelligence, University of Chinese Academy of Sciences\\[1pt]
$^4$Alibaba Group \quad $^5$Northwestern University \quad $^6$Xi'an Jiaotong University\\[1pt]
$^7$PeopleAI Inc. \quad $^8$ShanghaiTech University}\\[3pt]
{\small\ttfamily
\href{mailto:caizijian2024@ia.ac.cn}{caizijian2024@ia.ac.cn}\quad
\href{mailto:cfyuan@nlpr.ia.ac.cn}{cfyuan@nlpr.ia.ac.cn}}\\[3pt]
{\footnotesize $^*$Equal contribution. \quad $^\dagger$Corresponding author.}
\end{minipage}%
}

\iclrfinalcopy
\hypersetup{
  hidelinks,
  pdftitle={HyWe: Unifying Search and Operation for Real-World Hybrid Web Tasks},
  pdfauthor={Zijian Cai, Haowei Liu, Kailai Chen, Lin Peng, Cong Wan, Ziqi Zhang, Chunfeng Yuan, Bing Li, Weiming Hu, Haiyang Xu}
}
\begin{document}

\maketitle
\lhead{Preprint} 
\suppressfloats[t] 

\begin{abstract}
Real-world hybrid web tasks need both information gathering and page
interaction, yet search-based and operation-based agents are developed
and evaluated separately. To bridge this gap, we introduce \framework,
a framework that unifies both capabilities within one model and trajectory.
Its two modes use dedicated prompts and tool sets to sustain sub-goal
execution; explicit switching selects the needed capability, and shared
history lets each mode build on the other's findings. To provide demonstrations of
these behaviors, we develop \datapipe, a fully automatic pipeline that
uses task prototypes, four switching patterns, and multi-agent synthesis
to produce verified hybrid trajectories. We train \agent on these
demonstrations to unify efficient information gathering and browser
interaction in a single 35B web agent. To evaluate hybrid task completion,
we introduce \benchmark: $100$ human-written tasks on websites distinct
from training, with human-written sub-goal checklists and an outcome-based
evaluation pipeline. Hybrid training enables \agent to achieve $47.0\%$
success on this benchmark, improving on the untrained base model by
$19.0$ percentage points and outperforming existing search-based and
operation-based agents. Even before training, the same base model solves
$28.0\%$ of tasks with \framework, versus $11.0\%$ with all tools
under one prompt.
Beyond hybrid tasks, \agent achieves $59.0\%$ and $88.3\%$ success on
Online-Mind2Web and WebVoyager, outperforming existing open-source
baselines and cutting action cost per successful task by $48.3\%$ and
$51.9\%$ compared with Fara-7B. Code, models, and data are available at
\url{https://github.com/czjdsg/HyWe}.
\end{abstract}

\section{Introduction}
\label{sec:intro}

\begin{figure}[t]
\centering
\includegraphics[width=\textwidth]{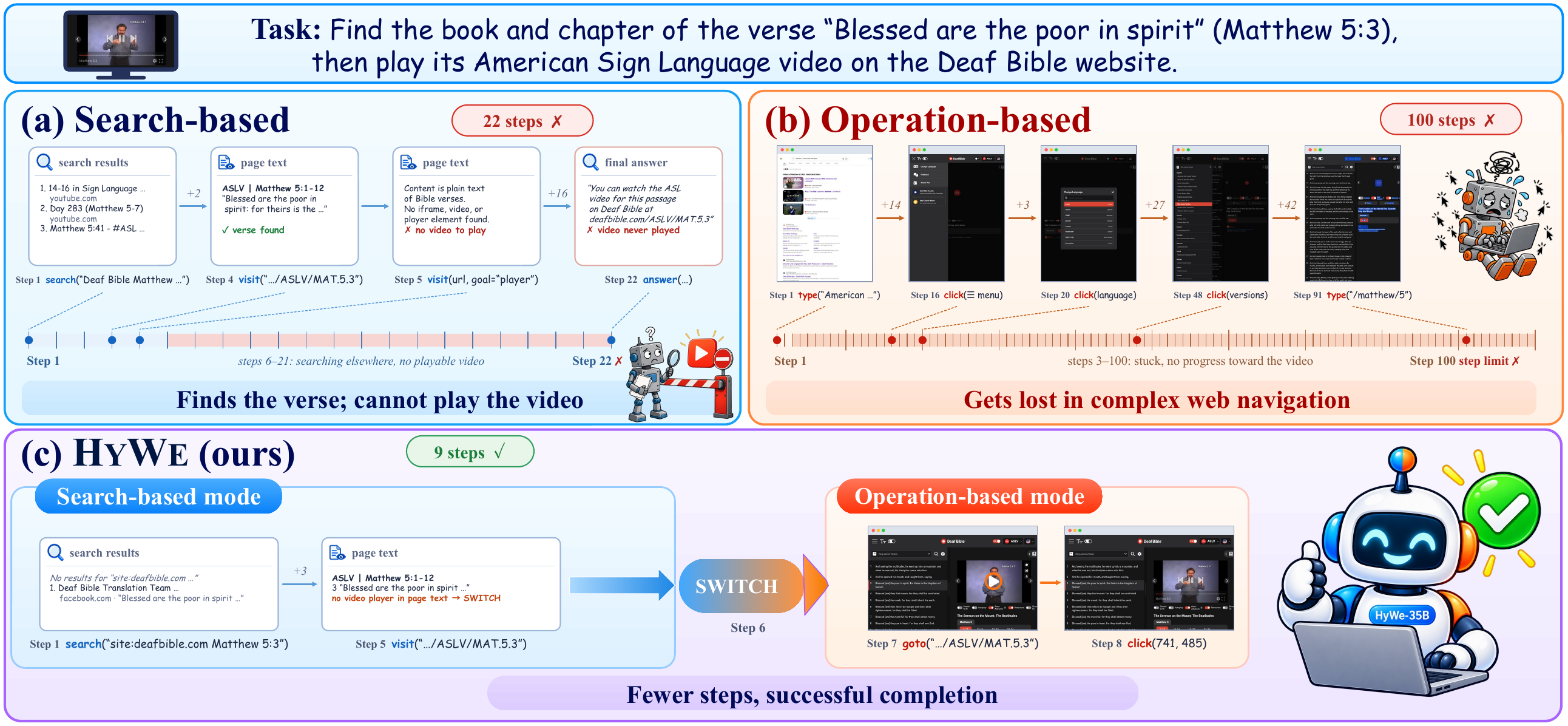}
\caption{One real-world hybrid task from \benchmark, three agents; key steps
are taken from their actual trajectories. (a)~A search agent finds the verse but cannot play the video.
(b)~An operation agent gets lost in complex web
navigation and reaches the 100-step limit. (c)~\agent finds the verse with
search tools, switches to operation tools, and plays the video in
9 steps.}
\label{fig:teaser}
\end{figure}

LLM agents have made substantial progress on web tasks, with approaches
broadly falling into two groups. (i)~Search agents use search
and page-reading APIs to gather information efficiently across websites
\citep{openai2025deepresearch,tongyi2025deepresearch}.
(ii)~Operation agents observe HTML or screenshots and
interact with browsers through clicking and typing
\citep{he2024webvoyager,qin2025uitars}.
In practice, however, many real-world web tasks are hybrid, and
both types of agents struggle to solve them. Figure~\ref{fig:teaser}
illustrates a task that requires both finding the Bible verse beginning
\task{Blessed are the poor in spirit} and playing its American Sign
Language video on Deaf Bible. Search agents quickly find the verse
but cannot play the video because their tools lack page interaction;
operation agents face long browser navigation to locate it.
Existing benchmarks also separate these task types: BrowseComp
\citep{wei2025browsecomp} and DeepSearchQA \citep{gupta2026deepsearchqa}
focus on cross-website information gathering, while Online-Mind2Web
\citep{xue2025onlinemind2web} and WebVoyager \citep{he2024webvoyager}
typically focus on operations within one website, leaving hybrid task
completion insufficiently evaluated.

To quantify the limitations of existing search and operation
agents, we sample real-world tasks from InSTA \citep{trabucco2025insta}
and probe capability and efficiency. For capability, the search
agent recovers only about $40\%$ of facts obtained through browser
interaction (\Secref{sec:method-data}). For efficiency, the search
agent solves $29\%$ of the sampled tasks within four steps, compared
with $6\%$ for the operation agent (Appendix~\ref{app:pilot}).
\textbf{Search agents gather information faster; operation agents use
browser actions to achieve required page states or obtain information
search cannot provide.}

To unify these complementary capabilities, we introduce \framework,
a framework for general-purpose web agents that organizes execution
around completing sub-goals. An agent should gather enough evidence or
complete the required interaction for each sub-goal, while using earlier
findings to guide the remaining work.
The same model works in a search mode or an operation mode.
Mode-specific prompts and isolated tool sets guide evidence gathering,
page interaction, and recovery from failed attempts, helping the model
focus on the current sub-goal. An explicit \textsc{switch} action makes
changing capabilities a deliberate decision: another mode may be needed
to finish the current sub-goal or pursue the next one more efficiently.
A unified action--observation format and shared history carry
earlier findings forward so subsequent work builds on existing progress.
Together, these designs help the agent complete sub-goals and choose
when to change modes. Compared to exposing all tools under one prompt,
our framework improves information and operation sub-goal completion
from $36.9\%$ and $15.8\%$ to $74.8\%$ and $43.7\%$, respectively,
using the same base model
(\Secref{sec:exp-ablation}).

To learn how to complete sub-goals and switch between modes, we develop
\datapipe, a fully automatic pipeline for collecting high-quality hybrid
trajectories on real websites. (i)~Task prototypes describe reusable
sub-goals; (ii)~four switching patterns capture capability limits,
efficiency, and dependencies between sub-goals; and (iii)~multi-agent
trajectory synthesis coordinates task generation, execution, phase
control, and review to produce verified examples of hybrid trajectories.
We use these examples to train \agent. As a result, it achieves higher success
rates and greater efficiency on
operation tasks: $59.0\%$ and $88.3\%$ success on Online-Mind2Web
and WebVoyager, using $48.3\%$ and $51.9\%$ fewer actions per successful
task than Fara-7B \citep{awadallah2025fara}. It also maintains strong deep
search capabilities, with $64.0\%$ accuracy on xbench-DeepSearch-2510
(\Secref{sec:exp-search}).

We introduce \benchmark to evaluate real-world hybrid task completion:
$100$ high-quality, human-written tasks on websites distinct
from those used for training,
human-written \emph{info}/\emph{op} checklists, and a corresponding
evaluation pipeline. Each task includes at least one \emph{op} sub-goal;
the judge verifies outcomes from execution evidence regardless of the
mode used. \agent achieves $47.0\%$ success, up from $28.0\%$ before
training, and outperforms search and operation agents,
including XYZ-Aquila-mini ($22.0\%$) and Fara-7B ($13.0\%$). Its success
rate is comparable to Kimi-K3 ($46.0\%$) and Qwen3.8-Max ($44.0\%$), with
$1/80$ and about $1/69$ of their total parameters, respectively
(\Secref{sec:exp-hybrid}).

Our contributions are as follows.

\textbf{(i)}~\textbf{\framework.} A framework unifying search and
operation capabilities through focused sub-goal execution,
explicit mode switching, and shared evidence.

\textbf{(ii)}~\textbf{\datapipe.} A fully automatic pipeline using task
prototypes, switching patterns, and multi-agent synthesis to collect
high-quality examples of hybrid trajectories.

\textbf{(iii)}~\textbf{\benchmark.} Human-written hybrid tasks and
\emph{info}/\emph{op} checklists, with an evaluation pipeline.

\textbf{(iv)}~\textbf{\agent.} Higher success and efficiency on
operation tasks, and substantially better hybrid task performance
than existing search and operation agents.

\section{Related Work}
\label{sec:related}

\paragraph{Search-based agents and benchmarks.}
Search-o1 \citep{li2025searcho1} and WebThinker \citep{li2025webthinker}
integrate web search into multi-step reasoning. Search-R1
\citep{jin2025searchr1} and DeepResearcher \citep{zheng2025deepresearcher}
train search use through reinforcement learning, while WebExplorer
\citep{liu2025webexplorer} and Tongyi DeepResearch
\citep{tongyi2025deepresearch} scale training with synthetic
information-seeking data. These methods focus on finding and reasoning
over web information. \framework unifies search-based tools with
operation-based actions and observations in one trajectory, allowing the
model to also obtain information produced by page interaction. For example,
it can use retrieved requirements to configure a product and read the
resulting price. \datapipe
provides demonstrations of when to choose each mode and how to pass
information between them.

Benchmarks used to evaluate search-based agents cover several aspects of
information gathering.
GAIA \citep{mialon2023gaia} evaluates tool-assisted reasoning,
BrowseComp \citep{wei2025browsecomp} targets hard-to-find facts, and
SealQA \citep{pham2025sealqa} tests reasoning with noisy or conflicting
search results. DeepSearchQA \citep{gupta2026deepsearchqa} further measures
whether agents retrieve complete answer sets. These benchmarks mainly
grade final answers against reference answers, which does not establish
whether an agent performed the page actions needed for a hybrid task.
\benchmark provides human-written tasks on real websites where retrieved information
guides page actions, or interaction results guide further search. Its
human-written checklists label information-gathering goals as \emph{info}
and operation goals as \emph{op}, with at least one \emph{op} sub-goal in every
task. Our evaluator checks both the facts obtained and the actions completed.

\paragraph{Operation-based agents and benchmarks.}
SeeAct \citep{zheng2024seeact} uses HTML and screenshots to ground browser
actions, while UI-TARS \citep{qin2025uitars} learns screenshot-based
interaction directly. UI-TARS-2 \citep{wang2025uitars2} further adds terminal
and file-system tools. OpenCUA \citep{wang2025opencua} scales training with
human demonstrations, while InSTA \citep{trabucco2025insta} and NNetNav
\citep{murty2024nnetnav} generate browser trajectories through website
interaction. Our method addresses the cost of gathering information
through repeated browser navigation. \framework lets the model use
search-based tools to gather information across websites and
operation-based tools for page interaction, with a shared history.
A single search can then replace several site visits when the required
facts are searchable.
\datapipe trains these mode choices through four switching patterns
grounded in capability probes.

Operation-based benchmarks differ in their environments and evaluation.
Mind2Web \citep{deng2023mind2web} evaluates action prediction against
recorded demonstrations, while WebArena \citep{zhou2023webarena} checks
task completion on self-hosted websites. Online-Mind2Web
\citep{xue2025onlinemind2web} and WebVoyager \citep{he2024webvoyager}
provide tasks on live websites and use models to judge browser
trajectories. Hybrid evaluation also needs evidence from search-based
tools, which may never appear in browser screenshots. \benchmark tests
tasks that require both information gathering and page interaction on
real websites. Our evaluator uses text tool observations and screenshots to
verify each sub-goal in the human-written checklists. This supports agents with
different tool sets and distinguishes completed work from unsupported
claims in the final answer.

\section{Method}
\label{sec:method}

Our method connects sub-goal execution, mode selection, and evidence
transfer within one trajectory (Figure~\ref{fig:overview}).
\framework includes mode-specific execution, mode switching, and shared
context (\Secref{sec:method-harness}); \datapipe collects
demonstrations for this framework (\Secref{sec:method-data}); and supervised
fine-tuning produces \agent (\Secref{sec:method-training}).

\subsection{HyWe framework}
\label{sec:method-harness}

\begin{figure}[!t]
\centering
\includegraphics[width=\textwidth]{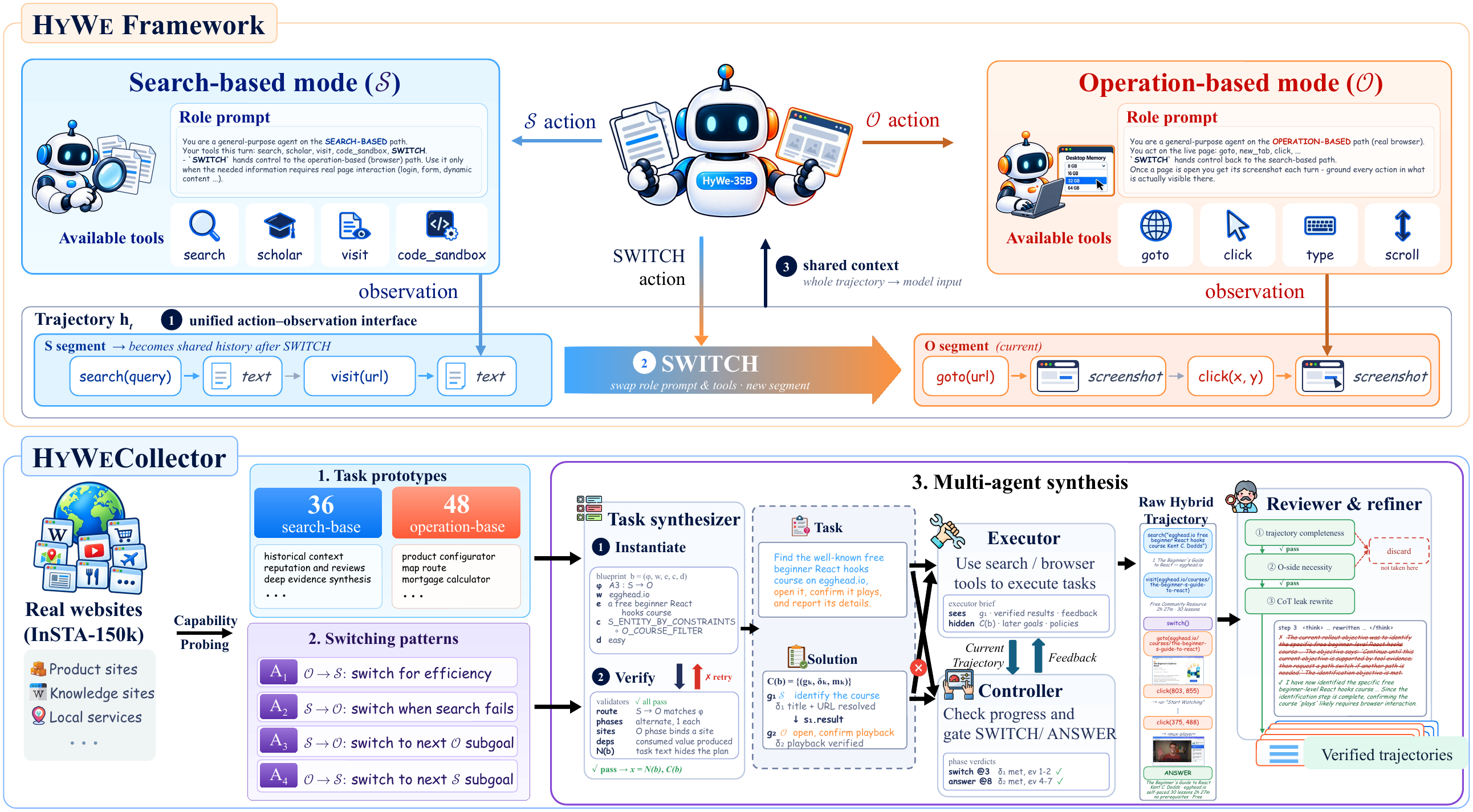}
\caption{Overview of the \framework framework.
Mode-specific prompts and tools guide sub-goal execution;
the \textsc{switch} action lets the model move between search-based and
operation-based modes as needed,
and shared history carries evidence forward (\Secref{sec:method-harness}).
The \datapipe data pipeline supplies task prototypes, switching patterns,
and website/entity bindings to a multi-agent framework for task synthesis,
execution, control, and review (\Secref{sec:method-data}). Colors mark
segments: consecutive steps in one mode, separated by mode switches.
Prompts and tool calls are abbreviated.}
\label{fig:overview}
\end{figure}

To sustain progress on hybrid tasks, \framework
(Figure~\ref{fig:overview}) guides execution within each mode,
lets the model switch between search-based and operation-based modes
as needed, and carries evidence across modes. Building on ReAct
\citep{yao2022react}, one model reasons
over observed progress to continue, switch, or answer
(Algorithm~\ref{alg:loop}).

\paragraph{Formulation.}
A sub-goal specifies a required fact or page outcome. A segment groups
consecutive steps in one mode; each executed \textsc{switch} ends the
current segment and starts the next. A sub-goal can span segments, and a
segment can advance several sub-goals. For task $x$, model $\pi_\theta$
produces trajectory $\tau=(o_0,a_1,o_1,\ldots,a_T)$, choosing action $a_t$
in mode $m_t\in\{\modeS,\modeO\}$ and receiving observation $o_t$:
\begin{align}
\mathcal{A}(m_t)&=\mathcal{A}_{m_t}\cup\{\textsc{switch},\textsc{answer}(y)\},
\label{eq:action-space}\\
a_t&\sim\pi_\theta\big(\cdot\mid\mathcal{R}(h_t,m_t)\big),\qquad
h_t=(x,(m_i,a_i,o_i)_{i<t}).
\label{eq:policy}
\end{align}
Here $\mathcal{A}_{m_t}$ contains the active mode's tools
(Appendix~\ref{app:harness}, Table~\ref{tab:tool-sig}), $h_t$ is the raw
history, and $\mathcal{R}$ renders the model's context.

\paragraph{Mode-specific execution.}
Each mode exposes its own prompt and tools to focus execution on the
current sub-goal. Search-based guidance asks the model to verify facts
from source pages and revisit unchecked constraints; operation-based
guidance asks it to verify page changes and recover from failed actions.
Both prompts require assessing remaining work and checking evidence
before answering (Appendix~\ref{app:prompts}). A common step format
$(m_t,a_t,o_t)$ lets the model reason over text search results and browser
screenshots within one trajectory.

\paragraph{Explicit mode switching.}
The model can switch when another mode supplies a missing capability or
advances a sub-goal more efficiently, including when earlier results enable
a dependent sub-goal. It states the reason and calls the argument-free
\textsc{switch} action, which changes $m_t$ to $\bar m_t$, activates the
other prompt and tool set, and starts a new segment without an environment
action. At each step, the model decides whether to continue in the current
mode, switch modes, or answer.

\paragraph{Shared context across modes.}
Later sub-goals may depend on entities, constraints, or results obtained
earlier. To preserve this evidence, steps preceding the current segment
are stored in a shared history block $H_{<s(t)}$, where $s(t)$ starts the
current segment. The model's context is
\begin{equation}
\mathcal{R}(h_t,m_t)=\Big[S_{m_t}\ ;\ U\big(x,H_{<s(t)}\big)\ ;\
(a_i,o_i)_{i=s(t)}^{t-1}\Big].
\label{eq:render}
\end{equation}
Here $S_{m_t}$ is the mode-specific prompt and $U$ holds the task and
history. On a switch, completed steps enter the history block with their
modes, actions, reasoning, and observations. The next segment thus uses
earlier evidence to assess unfinished work and continue from existing
progress. Context compression and screenshot handling are detailed in
Appendix~\ref{app:compression}.

\begin{algorithm}[t]
\caption{\framework inference loop}
\label{alg:loop}
\begin{algorithmic}[1]
\Require task $x$; initial mode $m_1$ (start URL if $m_1=\modeO$); step budget $T_{\max}$
\State $h_1\gets(x)$;\quad $s(1)\gets1$
\For{$t=1,\dots,T_{\max}$}
  \State $c_t\gets\mathcal{R}(h_t,m_t)$ \Comment{render context: Eq.~(\ref{eq:render})}
  \State $(r_t,a_t)\sim\pi_\theta(\cdot\mid c_t)$,\quad $a_t\in\mathcal{A}(m_t)$ \Comment{assess progress; choose one action}
  \State \textbf{if} $a_t=\textsc{answer}(y)$ \textbf{then return} $y$
  \If{$a_t=\textsc{switch}$}
     \State $o_t\gets$ hand-off notice;\quad $m_{t+1}\gets\bar m_t$;\quad $s(t{+}1)\gets t{+}1$
  \Else
     \State $o_t\gets\mathrm{Env}_{m_t}(a_t)$;\quad $m_{t+1}\gets m_t$;\quad $s(t{+}1)\gets s(t)$
  \EndIf
  \State $h_{t+1}\gets h_t\oplus(m_t,a_t,o_t)$ \Comment{append the observed step}
\EndFor
\State \Return forced answer \Comment{only \textsc{answer} is admitted at the budget limit}
\end{algorithmic}
\end{algorithm}

\subsection{HyWeCollector trajectory synthesis pipeline}
\label{sec:method-data}

To our knowledge, no open-source work has unified search-based and
operation-based trajectories for web tasks. We develop \datapipe
(Figure~\ref{fig:overview}) to collect demonstrations for \framework
through three fully automatic stages: abstracting reusable sub-goals,
designing switching patterns, and coordinating agents to synthesize
tasks and verified trajectories on real websites.

\paragraph{Stage 1: task prototypes from capability probing.}
To identify sub-goals suited to each mode, an operation-based exploration
agent visits InSTA websites \citep{trabucco2025insta} and an extractor
records facts such as prices and filtered counts. A search-based agent
tries to recover them within budget $B$. For the stable, reliable fact
pool $F$, a judge defines
\begin{equation}
\ell(f)=\mathbf{1}\big[\mathrm{Probe}_{\modeS}(f;B)\ \text{recovers}\ f\big],
\quad
F_{\modeS}=\{f\in F:\ell(f)=1\},\quad F_{\modeO}=F\setminus F_{\modeS}.
\label{eq:probe}
\end{equation}
Of all collected facts, $40\%$ enter $F_{\modeS}$ and $31\%$ enter
$F_{\modeO}$: $18\%$ depend on interaction and $13\%$ reflect probe limits.
From $F_{\modeS}$ and $F_{\modeO}$, we abstract
$36$ search-based and $48$ operation-based task prototypes
(Appendix~\ref{app:prototypes}), such as looking up requirements or
configuring a product. Each prototype specifies a sub-goal and typed
inputs and outputs for linking it to later sub-goals. Website and entity
bindings ground these prototypes during task synthesis.

\paragraph{Stage 2: switching pattern design.}
Our pilot study (Appendix~\ref{app:pilot}) shows that search-based agents
gather information faster, with low token costs even on failed attempts,
while operation-based agents
can solve tasks that search-based agents cannot. This motivates two
patterns that change capabilities while pursuing the same goal:
$\mathrm{A}_1$, $\modeO\!\to\!\modeS$ for efficiency, and
$\mathrm{A}_2$, $\modeS\!\to\!\modeO$ after a bounded search attempt fails.
Two further patterns connect dependent sub-goals: $\mathrm{A}_3$,
$\modeS\!\to\!\modeO$ with search results guiding page interactions, and
$\mathrm{A}_4$, $\modeO\!\to\!\modeS$ with interaction results guiding
further search. Stage~3 instantiates these patterns using the prototypes
and site bindings from Stage~1 (Appendix~\ref{app:archetypes}).

\paragraph{Stage 3: multi-agent trajectory synthesis.}
Four agent roles turn these sub-goals and switching patterns into
demonstrations of sustained execution and evidence transfer.

(i)~\emph{Task synthesizer.} It composes prototypes with
matching input/output types, binds them to real sites and entities,
and builds a task blueprint and private plan:
\begin{equation}
b=(\phi,w,e,c,d),\qquad
C(b)=\big((g_k,\delta_k,m_k)\big)_{k=1}^{K}.
\label{eq:contract}
\end{equation}
Here $\phi$ is the switching pattern, $w$ the target sites, $e$ the entity,
$c$ the prototype composition, and $d$ the difficulty. Phase $k$ specifies
goal $g_k$, condition $\delta_k$ for ending the phase, and mode $m_k$.
The synthesizer writes $x=N(b)$ without revealing the plan;
validators check site bindings and dependencies (Appendix~\ref{app:validators}).

(ii)~\emph{Executor.} It acts through \framework using the
current goal, verified earlier results, and feedback without plan details;
the full plan remains hidden.

(iii)~\emph{Controller.} It checks $\delta_k$ against observed
progress, approving a switch when the phase condition is met and an
answer after all phases finish. For $\mathrm{A}_2$, the condition is a
failed bounded search attempt. Execution and control are expressed as
\begin{equation}
a_t\sim\pi_E\big(\cdot\mid x,g_k,\beta,\sigma(u),\tau_{<t}\big),
\qquad v_k=\nu(\delta_k,\tau_k).
\label{eq:execution-control}
\end{equation}
Here $\pi_E$ is the executor, $\beta$ holds verified earlier results,
and $\sigma(u)$ removes plan details from controller feedback $u$.
The executor observes history $\tau_{<t}$; the controller $\nu$ evaluates
phase evidence $\tau_k$ and returns a binary verdict $v_k$.
Their outputs are a raw trajectory $\tau$ and a feedback record.

(iv)~\emph{Reviewer \& refiner.} This role verifies task completion
independently of the planned route, using the feedback record to identify
steps $L$ containing guidance leakage. Refinement rewrites these steps
without changing facts or decisions, followed by acceptance checks:
\begin{equation}
\widetilde{\tau}=\mathrm{Refine}(\tau,L),\qquad
\mathrm{Keep}(\widetilde{\tau})=
\mathrm{Success}(x,\widetilde{\tau})\land
\mathrm{Valid}(\widetilde{\tau},C(b)).
\label{eq:review-refine}
\end{equation}
$\mathrm{Success}$ checks task completion; $\mathrm{Valid}$ checks evidence
support, required interactions, switching conditions, and residual leakage.

In all, we probe facts from $1{,}110$
real websites and construct $35{,}784$ distinct tasks. We run the multi-agent
framework on $25{,}662$ of these tasks and retain $3{,}735$ trajectories after
review, refinement, and deduplication. Appendix~\ref{app:algorithms}
provides the full pipeline (Algorithm~\ref{alg:pipeline}) and detailed
executor--controller interaction (Algorithm~\ref{alg:rollout}).

\subsection{Training}
\label{sec:method-training}

\paragraph{Hybrid training samples.}
We convert the $3{,}735$ trajectories collected by \datapipe into $8{,}055$
multi-turn training samples. We split each trajectory $\tau$ at executed
switches and construct one sample per segment using the context renderer
$\mathcal{R}$ in Eq.~(\ref{eq:render}). At each supervised step $t$, the model
receives $\mathcal{R}(h_t,m_t)$, with earlier segments stored in the history
block $H_{<s(t)}$; only assistant turns in the current segment are supervised.
Thus, each trajectory yields
\begin{equation}
|\mathcal{D}(\tau)|=n_{\mathrm{sw}}(\tau)+1,
\label{eq:export}
\end{equation}
where $\mathcal{D}(\tau)$ is the set of exported training samples and
$n_{\mathrm{sw}}(\tau)$ the number of executed switches. This jointly
supervises work within a mode and decisions to \textsc{switch} or
\textsc{answer}, conditioned on evidence from earlier segments.

\paragraph{Search-based training data.}
To strengthen search-based capabilities, we supplement the hybrid samples
with the $7{,}000$ public search-based trajectories of \citet{xyz2026ai4ai},
adapting their tool names, observations, and prompts to \framework.
The hybrid trajectories already contain operation-based interactions, so
we use no additional operation-based training corpus.

Both sources use the inference context format. We fine-tune the
language-model parameters of Qwen3.6-35B-A3B \citep{qwen36_35b_a3b} with
next-token cross-entropy, keeping the vision encoder and projector frozen.
Sample processing and training settings are in Appendix~\ref{app:training}.

\section{Experiments}
\label{sec:exp}

\subsection{Setup}
\label{sec:exp-setup}

\paragraph{Benchmarks.}
We evaluate operation-based success and efficiency on Online-Mind2Web
\citep{xue2025onlinemind2web} and WebVoyager \citep{he2024webvoyager},
hybrid task completion on \benchmark, and deep search on HLE
\citep{phan2025hle}, xbench-DeepSearch-2510 \citep{chen2025xbench}, and
BrowseComp-EN \citep{wei2025browsecomp} (Appendix~\ref{app:eval}).

\paragraph{Baselines.}
We compare with open-source search-based and operation-based agents and
published frontier results. On \benchmark, frontier models also run within
\framework. The Qwen3.6-35B-A3B base model \citep{qwen36_35b_a3b} provides
training, framework, and mode controls. Baseline details are in
Appendix~\ref{app:baselines}.

\subsection{Success and efficiency on operation-based benchmarks}
\label{sec:exp-main}

Table~\ref{tab:main} compares success, efficiency, and switching.
Baseline success rates are published results; efficiency is measured on
our re-runs (Appendix~\ref{app:operation-protocol}).

\begin{table}[!t]
\centering
\caption{Success and efficiency on Online-Mind2Web and WebVoyager.
SR: task success rate (\%); Steps/SR: total actions (including failures)
per success in our re-runs, using separate success labels from the SR column;
Switch: tasks with a mode switch (\%). Bold/underline: best/second-best
SR and Steps/SR among open-source and \framework rows.
Small values: changes from base (rates in percentage points);
green/red/gray: improvement/decline/switch-rate changes.
$\uparrow$/$\downarrow$: higher/lower is better; --: unreported.}

\label{tab:main}
\footnotesize
\setlength{\tabcolsep}{3pt}
\begin{tabular}{@{}l c ccc ccc@{}}
\toprule
\multirow{2}{*}{\textbf{Model / system}} & \multirow{2}{*}{\textbf{Size}} & \multicolumn{3}{c}{\textbf{Online-Mind2Web}} & \multicolumn{3}{c}{\textbf{WebVoyager}} \\
\cmidrule(lr){3-5} \cmidrule(lr){6-8}
& & \textbf{SR}$\uparrow$ & \textbf{Steps/SR}$\downarrow$ & \textbf{Switch} & \textbf{SR}$\uparrow$ & \textbf{Steps/SR}$\downarrow$ & \textbf{Switch} \\
\midrule
\rowcolor{groupbg}\multicolumn{8}{l}{\textsc{Frontier LLMs}} \\
GPT-5.5 & -- & 74.7 & -- & -- & 85.1 & -- & -- \\
GPT-5 (SoM) & -- & 57.7 & -- & -- & 90.6 & -- & -- \\
o3 (SoM) & -- & 55.4 & -- & -- & 79.3 & -- & -- \\
Claude-Opus-4.6 & -- & -- & -- & -- & 88.0 & -- & -- \\
Seed2.1-Pro & -- & 59.2 & -- & -- & 70.6 & -- & -- \\
GLM-5V-Turbo & -- & -- & -- & -- & 88.5 & -- & -- \\
Kimi-K2.6 & -- & -- & -- & -- & 76.8 & -- & -- \\
Qwen3.8-Max & 2.4T-A95B & 68.9 & -- & -- & 77.8 & -- & -- \\
Gemini 2.5 Computer Use & -- & 57.3 & -- & -- & -- & -- & -- \\
OpenAI Operator & -- & 58.3 & -- & -- & 87.0 & -- & -- \\
Yutori Navigator (n1) & -- & 64.7 & -- & -- & -- & -- & -- \\
\midrule
\rowcolor{groupbg}\multicolumn{8}{l}{\textsc{Opensource operation-based agents}} \\
Fara-7B & 7B & 34.1 & 61.5 & 0 & 73.5 & 34.9 & 0 \\
UI-TARS-1.5-7B & 7B & 31.3 & 417.1 & 0 & 66.4 & 194.1 & 0 \\
GUI-Owl-1.5-8B-Thinking & 8B & 48.6 & 207.5 & 0 & 78.1 & 87.3 & 0 \\
GUI-Owl-1.5-32B-Thinking & 32B & -- & -- & -- & \underline{82.1} & 59.5 & 0 \\
MolmoWeb-8B & 8B & 35.3 & 149.8 & 0 & 78.2 & 102.2 & 0 \\
\midrule
\rowcolor{groupbg}\multicolumn{8}{l}{\framework Framework} \\
Qwen3.6-35B-A3B (base) & 35B-A3B & \underline{51.7} & \textbf{30.9} & 37.3 & 76.8 & \underline{20.8} & 20.8 \\
\rowcolor{oursbg}\agent & 35B-A3B & \textbf{59.0}\dgain{+7.3} & \underline{31.8}\dloss{+0.9} & 93.7\dneu{+56.4} & \textbf{88.3}\dgain{+11.5} & \textbf{16.8}\dgain{$-$4.0} & 87.9\dneu{+67.1} \\
\bottomrule
\end{tabular}
\end{table}

\paragraph{Success and efficiency.}
\agent leads the evaluated open-source agents with $59.0\%$ and $88.3\%$
success on Online-Mind2Web and WebVoyager. Even before training, the base
model in \framework uses fewer actions per success than every listed
operation-based agent: $30.9$ and $20.8$ Steps/SR, versus at least $61.5$
and $34.9$. Training raises success by $7.3$ and $11.5$ points, lowers
WebVoyager Steps/SR to $16.8$, and keeps Online-Mind2Web efficiency
comparable ($31.8$). Thus, the success gains retain the framework's
advantage in execution cost, including actions spent on failed tasks.

\paragraph{How switching helps.}
Search avoids long browser navigation for retrievable facts ($\mathrm{A}_1$);
switching back completes interaction-dependent goals ($\mathrm{A}_2$).
All trajectories start in $\modeO$. On Online-Mind2Web/WebVoyager,
training raises switch rates from $37.3\%$/$20.8\%$ to $93.7\%$/$87.9\%$
and mean switches from $0.44$/$0.23$ to $2.01$/$1.18$.
Still, $79\%$/$95\%$ of \agent trajectories switch at most twice:
$48\%$/$71\%$ follow $\modeO\to\modeS$ and $25\%$/$12\%$ follow
$\modeO\to\modeS\to\modeO$. These patterns indicate focused use of search
and page interaction with limited back-and-forth.

\subsection{Hybrid capability}
\label{sec:exp-hybrid}

\paragraph{\benchmark.}
Our benchmark contains $100$ high-quality, human-written hybrid tasks on websites
distinct from training, paired with human-written completion
checklists. Items specify information to obtain (\emph{info}) or
interactions and page states to achieve (\emph{op}); every task includes
at least one \emph{op} item. Task success requires completing all items.


\paragraph{Evaluation protocol.}
An LLM judge verifies each item's outcome from text observations and
screenshots, regardless of the tools or mode used. All systems share the
same tasks, budget, and checklists, keeping evaluation fair across agent
types. Checklist statistics and scoring details are in
Appendix~\ref{app:hybrid-protocol}.

\paragraph{Comparison with existing single-mode agents.}
\agent achieves $47.0\%$ success, versus $22.0\%$ for XYZ-Aquila-mini
and $13.0\%$ for Fara-7B (Table~\ref{tab:hwb-single}), more than doubling
the best single-mode result. It also leads the single-mode agents on both
information ($82.2\%$) and operation ($68.8\%$) sub-goals, turning
progress on both sides into more complete hybrid solutions. Training
improves task and operation sub-goal completion by $19.0$ and $25.1$
points, respectively. Its success rate is comparable to Kimi-K3
($46.0\%$) and Qwen3.8-Max ($44.0\%$), with $1/80$ and about $1/69$
of their total parameters, respectively.

\paragraph{Why search-based agents receive operation credit.}
The \emph{op} labels specify required outcomes, not mandatory tools.
Search-based agents receive credit when HTTP requests, parameterized URLs,
or retrieved content provide evidence that the required outcome was achieved.

\begin{table}[!t]
\centering
\caption{Performance on \benchmark. $\modeS$/$\modeO$ denote search-based/
operation-based modes. Success: tasks with all checklist items achieved
(\%); Info.\ and Op.: achieved information and operation checklist items
(\%), respectively. Sw.: mean switches per task; \%Sw.: tasks with at least
one switch (\%). Bold/underlined values mark the best/second-best scores;
$\uparrow$: higher is better; --: unavailable or inapplicable.}
\label{tab:hwb-single}
\small
\setlength{\tabcolsep}{1.8pt}
\begin{tabular}{l c c c c c c c}
\toprule
\textbf{System} & \textbf{Size} & \textbf{Mode} & \textbf{Success}$\uparrow$ & \textbf{Info.}$\uparrow$ & \textbf{Op.}$\uparrow$ & \textbf{Sw.} & \textbf{\%Sw.} \\
\midrule
\rowcolor{groupbg}\multicolumn{8}{l}{\textsc{Search-based agents}} \\
XYZ-Aquila-mini \citep{xyz2026ai4ai} & 35B-A3B & $\modeS$ & 22.0 & 71.2 & 26.7 & -- & -- \\
Agents-A1 \citep{bai2026agentsa1}             & 35B-A3B & $\modeS$ & 13.0 & 70.9 & 23.1 & -- & -- \\
\midrule
\rowcolor{groupbg}\multicolumn{8}{l}{\textsc{Operation-based agents}} \\
GUI-Owl-1.5 \citep{xu2026mobileagentv35}      & 32B     & $\modeO$ & 7.0 & 34.3 & 20.2 & -- & -- \\
Fara-7B \citep{awadallah2025fara}             & 7B      & $\modeO$ & 13.0 & 37.5 & 35.2 & -- & -- \\
MolmoWeb \citep{gupta2026molmoweb}            & 8B      & $\modeO$ & 2.0 & 11.7 & 9.7 & -- & -- \\
UI-TARS-1.5 \citep{qin2025uitars}             & 7B      & $\modeO$ & 6.0 & 24.6 & 21.1 & -- & -- \\
\midrule
\rowcolor{groupbg}\multicolumn{8}{l}{\framework Framework} \\
\rowcolor{oursbg}Kimi-K3 (zero-shot)      & 2.8T    & $\modeS{+}\modeO$ & \underline{46.0} & \textbf{83.2} & \underline{64.4} & 0.57 & 57 \\
\rowcolor{oursbg}Qwen3.8-Max (zero-shot)  & 2.4T    & $\modeS{+}\modeO$ & 44.0 & 77.3 & 62.3 & 0.59 & 54 \\
\rowcolor{oursbg}Qwen3.7-Plus (zero-shot) & --      & $\modeS{+}\modeO$ & 33.0 & 77.0 & 51.8 & 0.44 & 42 \\
\midrule
\rowcolor{oursbg}Qwen3.6-35B-A3B (base) & 35B-A3B & $\modeS{+}\modeO$ & 28.0 & 74.8 & 43.7 & 0.47 & 40 \\
\rowcolor{oursbg}\agent (SFT)             & 35B-A3B & $\modeS{+}\modeO$ & \textbf{47.0} & \underline{82.2} & \textbf{68.8} & 2.10 & 99 \\
\bottomrule
\end{tabular}
\end{table}

\subsection{Framework and mode ablations}
\label{sec:exp-ablation}

Table~\ref{tab:hybrid} tests whether mode organization helps hybrid task
completion and what each mode contributes, using the same base model.

\begin{table}[!t]
\centering
\caption{Framework and mode ablations on \benchmark (all $100$ tasks).
Mode: available tools ($\modeS$ search-based, $\modeO$ operation-based).
Succ.: tasks with all sub-goals achieved (\%); Info./Op.: completion rates
of information/operation sub-goals (\%). Bold: best base-model scores
within each panel; \agent is a training reference.}

\label{tab:hybrid}
\footnotesize
\setlength{\tabcolsep}{4pt}
\begin{tabularx}{\textwidth}{@{}X c ccc@{}}
\toprule
\textbf{System} & \textbf{Mode} & \textbf{Succ.}$\uparrow$ & \textbf{Info.}$\uparrow$ & \textbf{Op.}$\uparrow$ \\
\midrule
\rowcolor{groupbg}\multicolumn{5}{l}{\textbf{(a) Framework design}} \\
Base model, naive merging            & $\modeS{+}\modeO$  & 11.0 & 36.9 & 15.8 \\
\rowcolor{oursbg}Base model, \framework & $\modeS{+}\modeO$ & \textbf{28.0} & \textbf{74.8} & \textbf{43.7} \\
\midrule
\rowcolor{groupbg}\multicolumn{5}{l}{\textbf{(b) Modes within \framework}} \\
Base model                            & $\modeS$           & 7.0 & 73.8 & 15.8 \\
Base model                            & $\modeO$           & 26.0 & 57.6 & \textbf{47.8} \\
Base model                            & $\modeS{+}\modeO$  & \textbf{28.0} & \textbf{74.8} & 43.7 \\
\midrule
\rowcolor{oursbg}\agent (SFT reference) & $\modeS{+}\modeO$ & 47.0 & 82.2 & 68.8 \\
\bottomrule
\end{tabularx}
\end{table}

\paragraph{Framework ablation.}
We compare \framework with \emph{naive merging}, which exposes search-based
and operation-based tools under one prompt (Table~\ref{tab:hybrid}, panel~a).
\framework raises success from $11.0\%$ to $28.0\%$. Information and operation
completion rise from $36.9\%$/$15.8\%$ to $74.8\%$/$43.7\%$.
These results show that our framework based on mode switching is more
effective than simply merging the search and operation action spaces.

\paragraph{Mode ablation within \framework.}
The modes have complementary strengths (panel~b): search-based mode
completes more information sub-goals, and operation-based mode more
operation sub-goals. Before training, using both modes raises success only
slightly over operation-based mode alone ($28.0\%$ vs.\ $26.0\%$).
Training with high-quality trajectories from \datapipe raises success to
$47.0\%$, substantially improving the model's ability to complete tasks
requiring both information gathering and page interaction.

\paragraph{Case study.}
In the reward-chart task (Appendix~\ref{app:case-study-hywebench}), \agent
finds the correct builder through search, then switches to browser
interaction and completes all seven sub-goals. The base model under the
same framework stops after gathering information; this contrast illustrates
how training helps turn retrieved evidence into completed page actions.

\subsection{Retention of deep search capabilities}
\label{sec:exp-search}

\begin{table}[!t]
\centering
\caption{Deep search performance (accuracy, \%; higher is better).
HLE: Humanity's Last Exam; DS-2510: xbench-DeepSearch-2510; BC-en:
BrowseComp-EN. Superscript $t$ marks text-only HLE results; others use the full set;
--: unreported.}
\label{tab:search-retention}
\footnotesize
\setlength{\tabcolsep}{4pt}
\begin{tabularx}{\textwidth}{@{}X c ccc@{}}
\toprule
\textbf{Model / system} & \textbf{Size} & \textbf{HLE}$\uparrow$ & \textbf{DS-2510}$\uparrow$ & \textbf{BC-en}$\uparrow$ \\
\midrule
\rowcolor{groupbg}\multicolumn{5}{l}{\textsc{Frontier LLMs}} \\
GPT-5.6 Sol & -- & 58.0 & -- & 90.4 \\
Claude Fable 5 & -- & 64.5 & -- & 88.0 \\
Qwen3.8-Max & 2.4T & 56.2 & -- & -- \\
Kimi-K3 & 2.8T & 56.0 & -- & 91.2 \\
Kimi-K2.6 & 1T & 54.0 & -- & 83.2 \\
GLM-5.2 & 744B & 54.7$^t$ & -- & -- \\
DeepSeek-V4-Pro (Preview) & 1.6T & 48.2$^t$ & -- & 83.4 \\
\midrule
\rowcolor{groupbg}\multicolumn{5}{l}{\textsc{Opensource search-based agents}} \\
Tongyi DeepResearch & 30B & 32.9$^t$ & 55.0 & 43.4 \\
OpenSeeker-v1 & 30B & -- & -- & 29.5 \\
REDSearcher & 30B & 34.3 & -- & 57.4 \\
MiroThinker-1.7-mini & 30B & 36.4$^t$ & 57.2 & 67.9 \\
FORT-Searcher & 30B & -- & 57.2 & 72.2 \\
QUEST-35B & 35B & 37.2$^t$ & -- & 64.6 \\
Agents-A1 & 35B & 47.6$^t$ & -- & 75.5 \\
XYZ-Aquila-mini & 35B & 51.1$^t$ & -- & 78.8 \\
\midrule
\rowcolor{groupbg}\multicolumn{5}{l}{\framework Framework} \\
Qwen3.6-35B-A3B (base) & 35B-A3B & 30.5 & 49.0 & 59.2 \\
\rowcolor{oursbg}\agent & 35B-A3B & 47.8 & 64.0 & 74.6 \\
\bottomrule
\end{tabularx}
\end{table}

Alongside higher success on operation-based and hybrid tasks, \agent
substantially improves over the base model on search tasks: from $30.5\%$
to $47.8\%$ on HLE, $49.0\%$ to $64.0\%$ on xbench-DeepSearch-2510,
and $59.2\%$ to $74.6\%$ on BrowseComp-EN (Table~\ref{tab:search-retention}).
The agent rarely calls \textsc{switch}, so these evaluations primarily
measure deep research in \framework's search mode
(Appendix~\ref{app:search-protocol}).

Its $64.0\%$ on xbench-DeepSearch-2510 is the highest among the reported
open-source baselines in our comparison. On BrowseComp-EN, \agent
($74.6\%$) is close to Agents-A1 ($75.5\%$) and above FORT-Searcher
($72.2\%$), showing strong search capability alongside its hybrid gains.

\section{Conclusion}

We presented \framework, \datapipe, and \benchmark to unify search-based
and operation-based capabilities, collect hybrid demonstrations, and
evaluate real-world hybrid task completion. The central design is to
sustain progress on sub-goals within each mode and carry evidence
forward when capabilities change. Framework ablations show that this
organization improves completion over exposing all tools together;
hybrid training further strengthens their joint use. The resulting
\agent outperforms the evaluated search-based and operation-based
agents on hybrid tasks, achieves higher success and efficiency on
operation-based tasks, and preserves deep search capabilities.
Our ablations assess the framework as a whole, and the capability
probe depends on its search tools and budget. Isolating individual
design contributions and improving robustness to changing websites
remain open directions.

\subsection*{AI use statement}

AI tools assisted with manuscript preparation, language polishing,
literature search, LaTeX formatting, and code debugging. Their use in
data synthesis and evaluation is described in \Secref{sec:method-data}
and Appendix~\ref{app:eval}. The authors take full responsibility for the
accuracy, originality, and integrity of the final manuscript and research
artifacts.

\subsection*{Ethics statement}

Our data pipeline and evaluations interact with live, publicly accessible
websites. Exploration, probing, and rollouts perform ordinary browsing
interactions and do not create accounts, make purchases, submit personal
information, or otherwise durably alter site state. We have checked the
collected data for personal information; the released data contains no
personal information. The fact pool contains public facts such as prices,
counts, and product attributes.
Web agents that can both research and operate websites are dual-use;
our release targets research reproducibility, and our operation-based mode
training data derives from audited synthetic hybrid
trajectories on public sites rather than from interactions with
authenticated or private services.

\subsection*{Reproducibility statement}

The inference harness---mode-specific tool sets, output protocol, and
segmented context management with exact folding watermarks---is specified in
\Secref{sec:method-harness}, with sampling parameters and seed derivation
given in Appendix~\ref{app:runtime}. The data pipeline, including the switching
patterns, naturalization validators, dual-blind phase control, audit
criteria, and the export invariant, is specified in \Secref{sec:method-data},
and training details in \Secref{sec:method-training} and
Appendix~\ref{app:training}. Appendix~\ref{app:eval} specifies benchmark
coverage, judge prompts, and scoring; all of our reported results use
full benchmark sets. Live websites can change,
so identical seeds do not guarantee identical observations. We release code and research artifacts, including the framework,
collection and evaluation pipelines, task blueprints, training data,
fine-tuned checkpoints, and the human-written \benchmark tasks and
info/op checklists.

\bibliography{references}

@article{yao2022react,
  title={{ReAct}: Synergizing Reasoning and Acting in Language Models},
  author={Yao, Shunyu and Zhao, Jeffrey and Yu, Dian and Du, Nan and Shafran, Izhak and Narasimhan, Karthik and Cao, Yuan},
  journal={arXiv preprint arXiv:2210.03629},
  year={2022}
}

@article{li2025searcho1,
  title={Search-o1: Agentic Search-Enhanced Large Reasoning Models},
  author={Li, Xiaoxi and Dong, Guanting and Jin, Jiajie and Zhang, Yuyao and Zhou, Yujia and Zhu, Yutao and Zhang, Peitian and Dou, Zhicheng},
  journal={arXiv preprint arXiv:2501.05366},
  year={2025}
}

@article{li2025webthinker,
  title={{WebThinker}: Empowering Large Reasoning Models with Deep Research Capability},
  author={Li, Xiaoxi and Jin, Jiajie and Dong, Guanting and Qian, Hongjin and Wu, Yongkang and Wen, Ji-Rong and Zhu, Yutao and Dou, Zhicheng},
  journal={arXiv preprint arXiv:2504.21776},
  year={2025}
}

@misc{openai2025deepresearch,
  title={Introducing Deep Research},
  author={{OpenAI}},
  howpublished={\url{https://openai.com/index/introducing-deep-research/}},
  year={2025}
}

@article{jin2025searchr1,
  title={{Search-R1}: Training {LLMs} to Reason and Leverage Search Engines with Reinforcement Learning},
  author={Jin, Bowen and Zeng, Hansi and Yue, Zhenrui and Yoon, Jinsung and Arik, Sercan and Wang, Dong and Zamani, Hamed and Han, Jiawei},
  journal={arXiv preprint arXiv:2503.09516},
  year={2025}
}

@article{zheng2025deepresearcher,
  title={{DeepResearcher}: Scaling Deep Research via Reinforcement Learning in Real-world Environments},
  author={Zheng, Yuxiang and Fu, Dayuan and Hu, Xiangkun and Cai, Xiaojie and Ye, Lyumanshan and Lu, Pengrui and Liu, Pengfei},
  journal={arXiv preprint arXiv:2504.03160},
  year={2025}
}

@article{tongyi2025deepresearch,
  title={Tongyi {DeepResearch} Technical Report},
  author={{Tongyi DeepResearch Team}},
  journal={arXiv preprint arXiv:2510.24701},
  year={2025}
}

@article{liu2025webexplorer,
  title={{WebExplorer}: Explore and Evolve for Training Long-Horizon Web Agents},
  author={Liu, Junteng and Li, Yunji and Zhang, Chi and Li, Jingyang and Chen, Aili and Ji, Ke and Cheng, Weiyu and Wu, Zijia and Du, Chengyu and Xu, Qidi and Song, Jiayuan and Zhu, Zhengmao and Chen, Wenhu and Zhao, Pengyu and He, Junxian},
  journal={arXiv preprint arXiv:2509.06501},
  year={2025}
}

@article{mialon2023gaia,
  title={{GAIA}: A Benchmark for General {AI} Assistants},
  author={Mialon, Gr{\'e}goire and Fourrier, Cl{\'e}mentine and Swift, Craig and Wolf, Thomas and LeCun, Yann and Scialom, Thomas},
  journal={arXiv preprint arXiv:2311.12983},
  year={2023}
}

@article{wei2025browsecomp,
  title={{BrowseComp}: A Simple Yet Challenging Benchmark for Browsing Agents},
  author={Wei, Jason and Sun, Zhiqing and Papay, Spencer and McKinney, Scott and Han, Jeffrey and Fulford, Isa and Chung, Hyung Won and Passos, Alex Tachard and Fedus, William and Glaese, Amelia},
  journal={arXiv preprint arXiv:2504.12516},
  year={2025}
}

@article{phan2025hle,
  title={Humanity's Last Exam},
  author={Phan, Long and Gatti, Alice and others},
  journal={arXiv preprint arXiv:2501.14249},
  year={2025}
}

@article{pham2025sealqa,
  title={{SealQA}: Raising the Bar for Reasoning in Search-Augmented Language Models},
  author={Pham, Thinh and Nguyen, Nguyen and Zunjare, Pratibha and Chen, Weiyuan and Tseng, Yu-Min and Vu, Tu},
  journal={arXiv preprint arXiv:2506.01062},
  year={2025}
}

@article{chen2025xbench,
  title={xbench: Tracking Agents Productivity Scaling with Profession-Aligned Real-World Evaluations},
  author={Chen, Kaiyuan and Ren, Yixin and Liu, Yang and Hu, Xiaobo and Tian, Haotong and Xie, Tianbao and Liu, Fangfu and others},
  journal={arXiv preprint arXiv:2506.13651},
  year={2025}
}

@inproceedings{zheng2024seeact,
  title={{GPT-4V(ision)} is a Generalist Web Agent, if Grounded},
  author={Zheng, Boyuan and Gou, Boyu and Kil, Jihyung and Sun, Huan and Su, Yu},
  booktitle={International Conference on Machine Learning (ICML)},
  year={2024},
  note={arXiv:2401.01614}
}

@inproceedings{he2024webvoyager,
  title={{WebVoyager}: Building an End-to-End Web Agent with Large Multimodal Models},
  author={He, Hongliang and Yao, Wenlin and Ma, Kaixin and Yu, Wenhao and Dai, Yong and Zhang, Hongming and Lan, Zhenzhong and Yu, Dong},
  booktitle={Annual Meeting of the Association for Computational Linguistics (ACL)},
  year={2024},
  note={arXiv:2401.13919}
}

@article{qin2025uitars,
  title={{UI-TARS}: Pioneering Automated {GUI} Interaction with Native Agents},
  author={Qin, Yujia and Ye, Yining and Fang, Junjie and Wang, Haoming and Liang, Shihao and others},
  journal={arXiv preprint arXiv:2501.12326},
  year={2025}
}

@article{wang2025uitars2,
  title={{UI-TARS-2} Technical Report: Advancing {GUI} Agent with Multi-Turn Reinforcement Learning},
  author={Wang, Haoming and Zou, Haoyang and Song, Huatong and others},
  journal={arXiv preprint arXiv:2509.02544},
  year={2025}
}

@article{awadallah2025fara,
  title={{Fara-7B}: An Efficient Agentic Model for Computer Use},
  author={Awadallah, Ahmed and Lara, Yash and Magazine, Raghav and Mozannar, Hussein and Nambi, Akshay and others},
  journal={arXiv preprint arXiv:2511.19663},
  year={2025}
}

@article{gupta2026molmoweb,
  title={{MolmoWeb}: Open Visual Web Agent and Open Data for the Open Web},
  author={Gupta, Tanmay and Wolters, Piper and Ma, Zixian and Sushko, Peter and Pang, Rock Yuren and others},
  journal={arXiv preprint arXiv:2604.08516},
  year={2026}
}

@article{trabucco2025insta,
  title={{InSTA}: Towards Internet-Scale Training For Agents},
  author={Trabucco, Brandon and Sigurdsson, Gunnar and Piramuthu, Robinson and Salakhutdinov, Ruslan},
  journal={arXiv preprint arXiv:2502.06776},
  year={2025}
}

@article{murty2024nnetnav,
  title={{NNetNav}: Unsupervised Learning of Browser Agents Through Environment Interaction in the Wild},
  author={Murty, Shikhar and Zhu, Hao and Bahdanau, Dzmitry and Manning, Christopher D},
  journal={arXiv preprint arXiv:2410.02907},
  year={2024}
}

@inproceedings{wang2025opencua,
  title={{OpenCUA}: Open Foundations for Computer-Use Agents},
  author={Wang, Xinyuan and Wang, Bowen and Lu, Dunjie and Yang, Junlin and Xie, Tianbao and others},
  booktitle={Advances in Neural Information Processing Systems (NeurIPS)},
  year={2025},
  note={arXiv:2508.09123}
}

@inproceedings{deng2023mind2web,
  title={{Mind2Web}: Towards a Generalist Agent for the Web},
  author={Deng, Xiang and Gu, Yu and Zheng, Boyuan and Chen, Shijie and Stevens, Samuel and Wang, Boshi and Sun, Huan and Su, Yu},
  booktitle={Advances in Neural Information Processing Systems (NeurIPS)},
  year={2023},
  note={arXiv:2306.06070}
}

@inproceedings{zhou2023webarena,
  title={{WebArena}: A Realistic Web Environment for Building Autonomous Agents},
  author={Zhou, Shuyan and Xu, Frank F and Zhu, Hao and Zhou, Xuhui and Lo, Robert and Sridhar, Abishek and Cheng, Xianyi and Ou, Tianyue and Bisk, Yonatan and Fried, Daniel and Alon, Uri and Neubig, Graham},
  booktitle={International Conference on Learning Representations (ICLR)},
  year={2024},
  note={arXiv:2307.13854}
}

@article{xue2025onlinemind2web,
  title={An Illusion of Progress? Assessing the Current State of Web Agents},
  author={Xue, Tianci and Qi, Weijian and Shi, Tianneng and Song, Chan Hee and Gou, Boyu and Song, Dawn and Sun, Huan and Su, Yu},
  journal={arXiv preprint arXiv:2504.01382},
  year={2025}
}

@article{gupta2026deepsearchqa,
  title={{DeepSearchQA}: Bridging the Comprehensiveness Gap for Deep Research Agents},
  author={Gupta, Nikita and Chatterjee, Riju and Haas, Lukas and Tao, Connie and Wang, Andrew and Liu, Chang and Oiwa, Hidekazu and Gribovskaya, Elena and Ackermann, Jan and Blitzer, John and Goldshtein, Sasha and Das, Dipanjan},
  journal={arXiv preprint arXiv:2601.20975},
  year={2026}
}

@misc{xyz2026ai4ai,
  title={{AI4AI} at Scale: A Full-Pipeline System for Enhancing {LLM} Agentic Capabilities},
  author={{XYZ Agentic Team}},
  year={2026},
  howpublished={\url{https://xyz-lab.ai/blogs/ai4ai-at-scale/}},
  note={Technical report, July 2026}
}

@article{du2026openseeker,
  title={{OpenSeeker}: Democratizing Frontier Search Agents by Fully Open-Sourcing Training Data},
  author={Du, Yuwen and Ye, Rui and Tang, Shuo and Zhu, Xinyu and Lu, Yijun and Cai, Yuzhu and Chen, Siheng},
  journal={arXiv preprint arXiv:2603.15594},
  year={2026}
}

@article{chu2026redsearcher,
  title={{REDSearcher}: A Scalable and Cost-Efficient Framework for Long-Horizon Search Agents},
  author={Chu, Zheng and Wang, Xiao and Hong, Jack and Fan, Huiming and Huang, Yuqi and Yang, Yue and Xu, Guohai and Zhao, Chenxiao and Xiang, Cheng and Hu, Shengchao and Kuang, Dongdong and Liu, Ming and Qin, Bing and Yu, Xing},
  journal={arXiv preprint arXiv:2602.14234},
  year={2026}
}

@article{bai2026agentsa1,
  title={Scaling the Horizon, Not the Parameters: Reaching Trillion-Parameter Performance with a {35B} Agent},
  author={Bai, Lei and Cao, Zongsheng and Chen, Yang and Cui, Zhiyao and Du, Shangheng and Fan, Yue and Feng, Shiyang and Guo, Zijie and He, Haonan and He, Liang and others},
  journal={arXiv preprint arXiv:2606.30616},
  year={2026}
}

@article{xu2026mobileagentv35,
  title={{Mobile-Agent-v3.5}: Multi-platform Fundamental {GUI} Agents},
  author={Xu, Haiyang and Zhang, Xi and Liu, Haowei and Wang, Junyang and Zhu, Zhaozai and Zhou, Shengjie and Hu, Xuhao and Gao, Feiyu and Cao, Junjie and others},
  journal={arXiv preprint arXiv:2602.16855},
  year={2026}
}

@article{xie2026quest,
  title={{QUEST}: Training Frontier Deep Research Agents with Fully Synthetic Tasks},
  author={Xie, Jian and Lin, Tianhe and Wang, Zilu and Ning, Yuting and Yao, Yuekun and Xue, Tianci and Zhang, Zhehao and Li, Zhongyang and Zhang, Kai and Wu, Yufan and others},
  journal={arXiv preprint arXiv:2605.24218},
  year={2026}
}

@misc{qwen36_35b_a3b,
    title = {{Qwen3.6-35B-A3B}: Agentic Coding Power, Now Open to All},
    url = {https://qwen.ai/blog?id=qwen3.6-35b-a3b},
    author = {{Qwen Team}},
    month = {April},
    year = {2026}
}

@misc{bytedance2025uitars15,
  title={{UI-TARS-1.5}: Open-Source Multimodal Agent},
  author={{ByteDance UI-TARS Team}},
  year={2025},
  howpublished={\url{https://github.com/bytedance/UI-TARS}},
  note={Model release, April 2025}
}

@article{miromind2026mirothinker17,
  title={{MiroThinker-1.7 \& H1}: Towards Heavy-Duty Research Agents via Verification},
  author={{MiroMind Team}},
  journal={arXiv preprint arXiv:2603.15726},
  year={2026}
}

@article{deng2026fortsearcher,
  title={{FORT-Searcher}: Synthesizing Shortcut-Resistant Search Tasks for Training Deep Search Agents},
  author={Deng, Jia and Chen, Yimeng and Xiang, Xiaoqing and Zeng, Ziyang and Tang, Shuo and Zhao, Wayne Xin and Chang, Feng and Hao, Chuan and Wei, Yuan and Tao, Ran and Dai, Bryan and Wen, Ji-Rong},
  journal={arXiv preprint arXiv:2606.12087},
  year={2026}
}
\bibliographystyle{iclr2027_conference}

\appendix
\clearpage

\section{Case studies}
\label{app:case-study}

\subsection{\benchmark}
\label{app:case-study-hywebench}

\begin{figure}[!htbp]
\centering
\includegraphics[width=\textwidth]{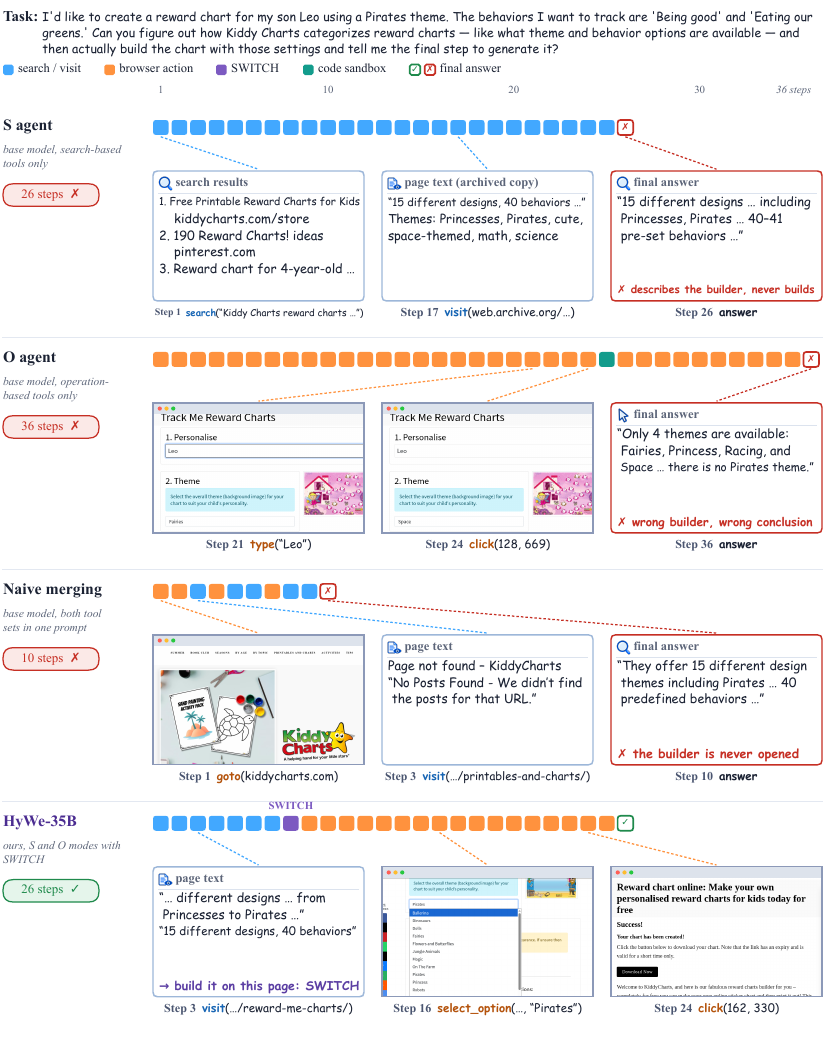}
\caption{Case study on one \benchmark task with seven sub-goals. Rows 1--3
use the same base model with search-based tools only, operation-based tools
only, or both tool sets under one prompt (naive merging); row 4 is \agent.
Each block of a strip is one step, and all strips share one step scale.
Dashed lines link key steps to their tool calls (bottom) and results (top).
Texts and screenshots are taken from the evaluation runs; quotations are
shortened.}
\label{fig:case-study}
\end{figure}

Figure~\ref{fig:case-study} compares four agents on the task \task{I'd like to
create a reward chart for my son Leo using a Pirates theme \ldots\ build the
chart with those settings and tell me the final step to generate it}. The
first three agents use the same base model, Qwen3.6-35B-A3B, and differ only
in their tools and prompt; the fourth is \agent.

\paragraph{Search-based agent.}
The chart builder is rendered by page scripts, so page reading returns only
its description (``15 different designs, 40 behaviors''). The agent spends 26
steps on searches, third-party reviews, and archived copies of the page, then
answers with a description of the builder. It never configures a chart and
achieves 4 of 7 sub-goals.

\paragraph{Operation-based agent.}
Without search-based tools, the agent navigates the site's menus and reaches a
different builder, Track Me Reward Charts, which offers four themes. After
entering the child's name and trying the theme menu, it concludes that no
Pirates theme exists. It achieves 5 of 7 sub-goals in 36 steps.

\paragraph{Naive merging.}
With both tool sets under one prompt, the agent alternates between browser
actions and search-based tools five times in ten steps. It opens the site in
the browser, reads a missing page through \texttt{visit}, scrolls the browser
page, and returns to search. After the first step, it outputs tool calls
without any reasoning. It never opens the builder, answers with a
description, and achieves 2 of 7 sub-goals.

\paragraph{\agent.}
Seven search-based steps establish that the Reward Me Charts builder offers a
Pirates theme and 40 behaviors. The agent then calls \textsc{switch}, stating
that the remaining details require interacting with the page, and completes
the chart in 17 browser actions: it enters the name, selects the Pirates theme,
checks both behaviors, and generates the chart. It achieves all 7 sub-goals in
26 steps. Under the same framework, the base model collects the same
information but never calls \textsc{switch} and achieves 3 of 7 sub-goals in
25 steps. This suggests that the gain comes from learning when to switch, not
from the tool interface alone.

\subsection{Operation benchmarks}
\label{app:case-study-op}

\begin{figure}[!htbp]
\centering
\includegraphics[width=\textwidth]{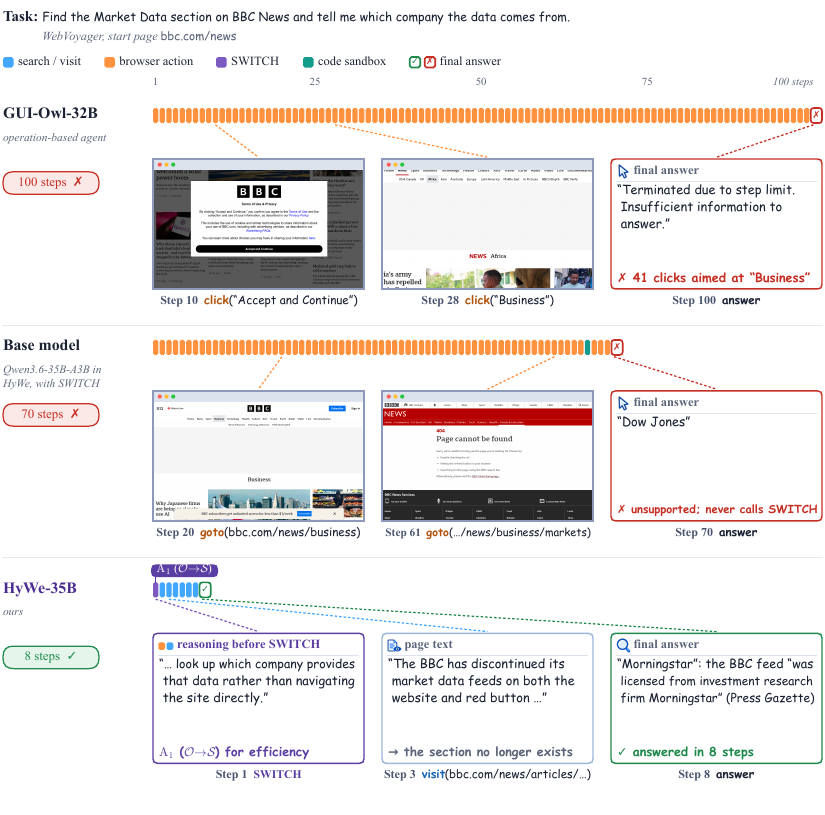}
\caption{Case study on a WebVoyager task, showing switching pattern
$\mathrm{A}_1$ ($\modeO\!\to\!\modeS$ for efficiency). The rows compare
GUI-Owl-32B, the base model under \framework, and \agent. The layout follows
Figure~\ref{fig:case-study}, and all strips share one step scale. The
GUI-Owl-32B screenshots come from its raw rollout.}
\label{fig:case-wv}
\end{figure}

\begin{figure}[!htbp]
\centering
\includegraphics[width=\textwidth]{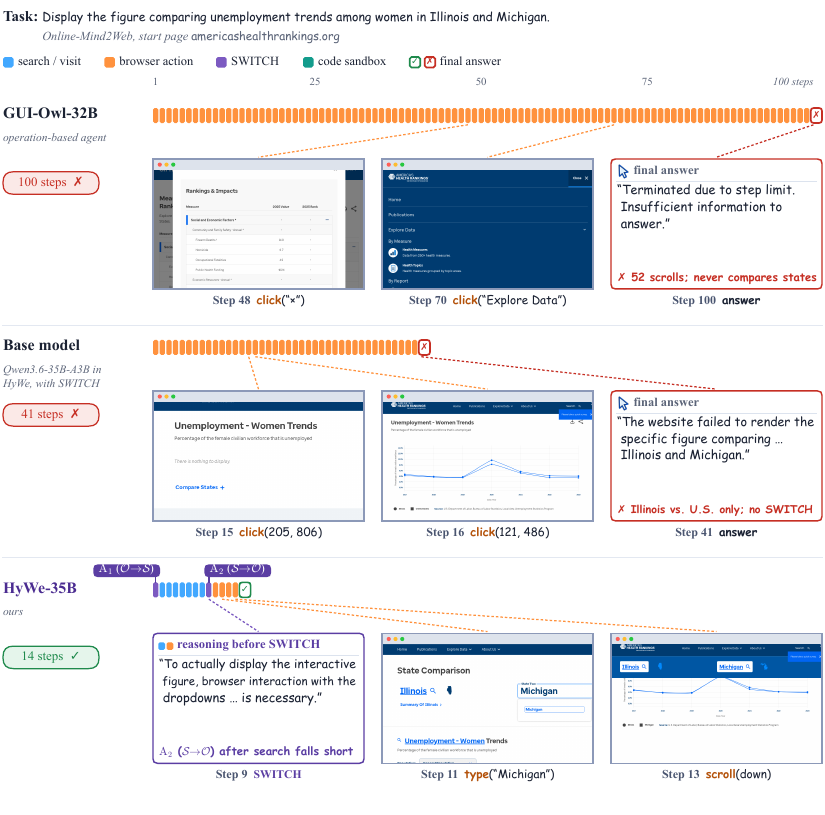}
\caption{Case study on an Online-Mind2Web task, showing switching patterns
$\mathrm{A}_1$ and $\mathrm{A}_2$ ($\modeS\!\to\!\modeO$ after a search
attempt falls short). The rows and layout are the same as in
Figure~\ref{fig:case-wv}.}
\label{fig:case-m2w}
\end{figure}

Figures~\ref{fig:case-wv} and~\ref{fig:case-m2w} show one task from
WebVoyager and one from Online-Mind2Web. Both tasks start from a web page.
Each figure compares \agent with GUI-Owl-32B and with the base model under
\framework.

\paragraph{$\mathrm{A}_1$ on WebVoyager.}
The task asks for the Market Data section of BBC News and the company that
supplies its data. The BBC discontinued this section in November 2024, so no
sequence of page actions can reach it. GUI-Owl-32B clicks the cookie dialog 11
times in its first 14 steps. Its next 41 clicks target the Business tab, but
none of them reaches the Business section, and it stops at the 100-step limit.
The base model under \framework browses the News and Business pages for 70
steps and opens a missing markets page. It never calls \textsc{switch} and
answers ``Dow Jones'', which no visited page supports. \agent calls
\textsc{switch} at its first step, stating that searching is faster than
navigating the site. Its searches find the BBC notice that the feeds were
discontinued and a Press Gazette report that the feed was licensed from
Morningstar. It answers correctly in 8 steps.

\paragraph{$\mathrm{A}_1$ and $\mathrm{A}_2$ on Online-Mind2Web.}
The task asks to display the chart that compares women's unemployment trends
in Illinois and Michigan on America's Health Rankings. GUI-Owl-32B scrolls 52
times and opens menus and ranking tables. It reaches the Compare States link
only at its 100th step, when the step limit ends the episode. The base model
reaches a chart of Illinois against the United States in 16 steps but does not
add Michigan. After 41 browser steps, it reports that the site failed to render
the chart. \agent first switches to search ($\mathrm{A}_1$) and finds the
state comparison page by step 7. Reading that page returns only the default
comparison of Illinois with the United States, because page reading cannot
operate the state selector. The agent then switches back to the browser
($\mathrm{A}_2$), opens the comparison page, selects Michigan, and displays the
chart, completing the task in 14 steps. In both tasks, the operation-based
baselines spend most of their step budget on navigation. \agent uses search to
locate the target and uses the browser only when the page must be operated.

\clearpage

\section{Pilot study: capability and efficiency}
\label{app:pilot}

\begin{figure}[!htb]
\centering
\includegraphics[width=\textwidth]{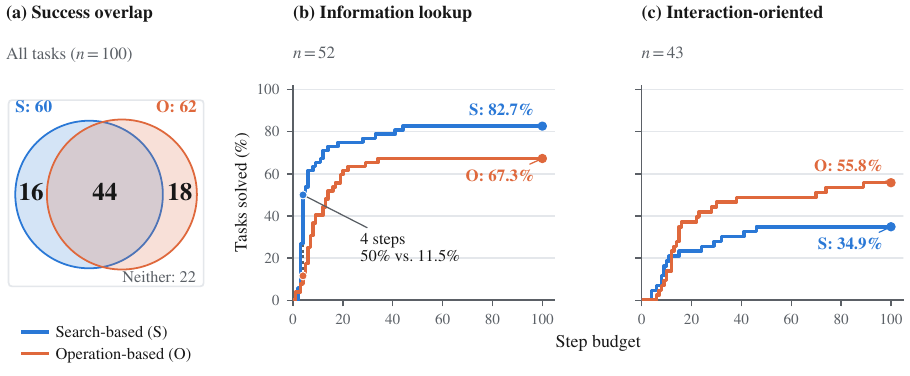}
\caption{Pilot study on $100$ InSTA tasks with search-based (S) and operation-based
(O) runs. \textbf{(a)}~Success overlap; circle and intersection areas are
proportional to success counts. \textbf{(b,c)}~Cumulative success rates for
information-lookup and interaction-oriented tasks, classified from task
descriptions and reference steps. Each denominator includes all tasks in its
stratum, including failures. The five uncertain tasks appear only in (a).
The curves use matching linear axes and count successful runs completed
within the indicated number of steps; they summarize fixed-budget runs,
not separate evaluations rerun at each budget.}
\label{fig:pilot}
\end{figure}

\paragraph{Evidence for $\mathrm{A}_1$ and $\mathrm{A}_2$.}
To motivate the two mode-selection patterns, we analyze $100$ tasks from InSTA
\citep{trabucco2025insta}, each run twice under the same harness and budget,
once with a search-based agent and once with an operation-based agent (Figure~\ref{fig:pilot}).

The two agents solve \emph{different tasks}. The search-based agent solves $60$
tasks and the operation-based agent $62$, but only $44$ are solved by both:
$16$ are solved only by the search-based agent and $18$ only by the
operation-based agent (Figure~\ref{fig:pilot}a).

Stratifying by \emph{task requirements} reveals different advantages.
Retrospective annotation of the original task descriptions and reference steps
identifies $52$ information-lookup tasks and $43$ interaction-oriented tasks,
such as conditional queries, configured quotes, and form submissions.
Five uncertain cases remain in the overlap analysis but are omitted from the
stratified curves. These labels describe task demands, not mode exclusivity.

The search-based agent has an \emph{early advantage on information lookup}: within four steps,
it solves $26/52$ tasks ($50.0\%$), compared with $6/52$ ($11.5\%$) for
the operation-based agent. Final success rates are $82.7\%$ and $67.3\%$, respectively
(Figure~\ref{fig:pilot}b).

The operation-based agent achieves \emph{higher coverage on interaction-oriented tasks},
solving $24/43$ ($55.8\%$), compared with $15/43$ ($34.9\%$) for the search-based agent
(Figure~\ref{fig:pilot}c). The search-based agent can still solve some of these tasks, so the
task category alone does not determine which mode is needed.

These observations support two design choices: \emph{prefer the search-based mode} when it
suffices ($\mathrm{A}_1$), and \emph{fall back to the operation-based mode} when the search-based mode fails
to obtain the required information ($\mathrm{A}_2$). When reachability is
uncertain, $\mathrm{A}_2$ permits a bounded search attempt before switching.
The pilot motivates these patterns; \Secref{sec:exp-hybrid} evaluates learned
switching.

\subsection{Cost and failure analysis}

\paragraph{Paired costs.}
The pilot contains $100$ tasks from $100$ domains. Cost ratios are computed
within each task and then summarized by their median; they are not ratios of
the two modes' median costs. Steps and token usage are available for all $44$
joint successes. Five of these operation-based runs lack wall-clock measurements,
so the time comparison uses the remaining $39$ pairs without imputation.
Two-sided Wilcoxon signed-rank tests on the paired raw costs give
$p=0.0089$ for steps, $p=5.15\times10^{-9}$ for tokens, and
$p=1.63\times10^{-9}$ for time. The median ratios of operation-based to search-based costs are
$1.9$, $9.0$, and $6.3$, respectively.

\paragraph{Tasks solved only by the search-based agent.}
Of the $16$ tasks solved only by the search-based agent, failure annotations
for the operation-based agent
identify incorrect or incomplete extraction in $7$, looping or giving up in
$5$, widget failures in $2$, and page-load failures in $2$. Thus $14$ failures
concern interpreting or interacting with pages rather than loading them.
The failed operation-based runs take a median of $25$ steps, compared with $10$
for their successful search-based counterparts. In $12$ of these $16$ tasks,
the search-based run bypasses on-site navigation through a deep link or a
parameterized URL.

\paragraph{Tasks solved only by the operation-based agent.}
Among the $18$ tasks solved only by the operation-based agent, failure
annotations for the search-based agent
identify inaccessible dynamic state in $7$, answers sourced only from other
sites in $5$, incorrect extraction in $4$, an incorrect claim of absence in
$1$, and another failure in $1$. The failed search-based run uses a median of
$0.22$ times the tokens of its successful operation-based counterpart. This measures
the overhead of these particular attempts; it does not determine an optimal
search budget. Successful operation-based runs are annotated as interaction-dependent
on $16/18$ tasks solved only by the operation-based agent and $26/44$ joint successes
(two-sided Fisher exact $p=0.0347$). Interaction dependence is a trajectory
annotation: a browser run may use a form even when search can obtain the answer
through a direct URL. The pilot's modes are locked, so these outcomes cannot
measure whether the model would choose to switch. Likewise, termination with
an answer action alone does not show whether its response acknowledges failure.

\subsection{Task selection and annotation}

\paragraph{Sample and task requirements.}
The $100$ tasks span $100$ domains and form a curated pilot rather than an
unfiltered random sample of InSTA. The selection script retains $81$ tasks
from the first candidate pool after excluding cases whose operation failure
is attributed primarily to anti-bot blocking, then samples $19$ tasks from a
second pool under the same exclusion rule (seed $20260915$). The results
therefore describe this filtered sample.

For Figure~\ref{fig:pilot}b,c, we retrospectively assign task-demand labels
using the original task wording and dataset reference steps. This is an
AI-assisted annotation, not an independently blinded or human-validated
label set. The annotation input records contain no paired outcomes, costs,
or rollout-level interaction attributions. Information lookup covers
retrieving, comparing, or summarizing existing page content; using a search
box or filters merely to locate a named product page does not suffice for
an interaction label. Interaction-oriented tasks request conditional query
results, filtered catalogue rankings or counts, configured prices,
service-generated outputs, or state-changing actions. These categories
describe requested outputs and actions; they do not establish that browser
operation is necessary. Search may reach a conditional result through a
parameterized URL or another permitted route.

Five tasks remain uncertain because the reference material does not resolve
whether the requested information is directly readable or depends on an
interface state: enquiry-form options, bike-station availability, shipment
availability for a part, size availability for a boot, and chapter-specific
audio languages. They are retained in all aggregate analyses and excluded
only from the two stratified panels. No fresh website reachability checks
are used to resolve these cases. Table~\ref{tab:pilot-requirements} reports
all three groups; success counts use every task in each group, including
operation-based failures. The task-level annotations retain
the original wording, reference steps, source identifiers, and a rationale
for each label.

\begin{table}[!htb]
\centering
\caption{Pilot results by task requirement. Successes are counts, with
success rates in parentheses. Each rate uses all tasks in its row as the
denominator, including failed runs.}
\label{tab:pilot-requirements}
\begin{tabular}{lrrr}
\toprule
Task demand & Tasks & Search-based successes & Operation-based successes \\
\midrule
Information lookup & $52$ & $43$ ($82.7\%$) & $35$ ($67.3\%$) \\
Interaction-oriented & $43$ & $15$ ($34.9\%$) & $24$ ($55.8\%$) \\
Uncertain & $5$ & $2$ ($40.0\%$) & $3$ ($60.0\%$) \\
\midrule
All tasks & $100$ & $60$ ($60.0\%$) & $62$ ($62.0\%$) \\
\bottomrule
\end{tabular}
\end{table}

\clearpage

\section{Evaluation and ablation details}
\label{app:eval}

\subsection{\benchmark tasks and evaluation}
\label{app:hybrid-protocol}
The benchmark's $100$ tasks and their checklists are written by humans.
Each checklist contains $2$--$11$ items ($5.6$ on average), including
$1$--$6$ \emph{op} items per task. The $556$ items comprise $309$
\emph{info} and $247$ \emph{op} sub-goals. Checklists are fixed before
evaluation and shared across all systems.

Each system runs every task once with a budget of $100$ steps. Tasks
with a start page begin on that page; the others start without a URL.
Kimi-K3 reads the complete execution record, including text tool outputs
and browser screenshots with their URLs and page titles. For each item,
it reports whether the requested outcome was achieved, the supporting
step, and the evidence. Labels describe the requested outcome and do
not prescribe tools or modes. External agents retain their native
prompts and action spaces; their execution frameworks are described in
Appendix~\ref{app:baseline-execution}.

Task success requires every checklist item to be achieved. Information
and operation completion rates aggregate achieved items of each type
across all $100$ tasks. Switching statistics report executed switches
per task and the fraction of tasks with at least one switch.
Successful-run length averages steps over successful tasks.

\subsection{Base-model and framework controls}
\label{app:controls}

\paragraph{Base model.}
Qwen3.6-35B-A3B \citep{qwen36_35b_a3b} is the checkpoint from which
\agent is fine-tuned. It has approximately 35B total parameters and
3B active parameters. We evaluate it within \framework before our
training and compare it with \agent under matched settings to measure
the effect of the training procedure in \Secref{sec:method-training}.

\paragraph{Framework and mode ablations.}
Using the same base model, naive merging exposes both tool sets under
one prompt without explicit modes or a switch action.
The full \framework configuration adds mode-specific instructions,
tool isolation, explicit switching, and shared history. We also evaluate
search-based-only and operation-based-only configurations within
\framework, removing \textsc{switch} and its prompt instructions in
each single-mode variant. Together, these controls distinguish the effect of
organizing tools into modes from the contribution of having access
to each tool set (Table~\ref{tab:hybrid}). The framework-design comparison
uses the same base model for both configurations on all $100$
tasks. Mode ablations use the same model on all $100$ tasks; the
\framework{} row in panel~(a) and the $\modeS{+}\modeO$ row in
panel~(b) are the same run.

\subsection{Execution frameworks on \benchmark}
\label{app:baseline-execution}

XYZ-Aquila-mini uses its released AxisAgentic search framework, and
Agents-A1 uses the search evaluation implementation in its released
repository. Fara-7B runs its official agent and browser environment;
GUI-Owl-1.5 uses the Mobile-Agent-v3.5 web evaluation implementation;
MolmoWeb-8B uses its official agent and benchmark runner. For
UI-TARS-1.5-7B, we implement the browser execution loop with Playwright,
using the released computer-use prompt and action parser. Trajectories
are converted to a common record format for the \benchmark judge.
The base model, \agent, frontier models, and internal mode ablations
use \framework, with the changes for each ablation described above.

\subsection{Operation-based evaluation protocol}
\label{app:operation-protocol}
Our agents start in operation-based mode and are evaluated on all $300$
Online-Mind2Web tasks and all $643$ WebVoyager tasks. We adapt WebJudge
\citep{xue2025onlinemind2web} and the WebVoyager auto-evaluator
\citep{he2024webvoyager} to include evidence from both modes, retaining
o4-mini for WebJudge and GPT-4o for the WebVoyager evaluator.

For the open-source operation-based baselines in Table~\ref{tab:main},
success rates are quoted from the original reports, while step counts
are measured on our re-runs judged with Kimi-K3. Published baselines may
use different evaluation settings. Steps/SR is the mean action count
over all retained task runs, including failures, divided by the success
rate of those same runs under Kimi-K3. Actions in both modes and switch
calls are counted. This denominator is separate from the published or
official-judge SR column. Switch reports the fraction of tasks with at
least one executed mode switch.

For example, Fara-7B uses only $12.2$ and $10.0$ actions on the tasks it
solves on Online-Mind2Web and WebVoyager. Counting actions from failed
runs as well raises its Steps/SR to $61.5$ and $34.9$, respectively,
about $2.0$ and $1.7$ times the base model's $30.9$ and $20.8$.

\paragraph{Adapting the operation-based benchmark judges to hybrid trajectories.}
\label{app:judge-adapt}
The official judges of Online-Mind2Web (WebJudge) and WebVoyager assess a
trajectory from its browser screenshots and final response. A hybrid
trajectory also contains steps in the search-based mode whose evidence is text---search
results, page contents read by \texttt{visit}, code outputs---and a
screenshot-only judge cannot see it. We make two changes and keep the official
prompts otherwise verbatim. (i)~\emph{WebJudge.} Each text observation is
rated against the task's key points on the same $1$--$5$ relevance scale and
with the same threshold ($3$) as screenshots; the selected observations enter
the final verdict next to the selected screenshots, introduced as page-content
evidence with the same status. (ii)~\emph{WebVoyager auto-evaluator.} Text
observations and screenshots are merged into one chronological timeline, from
which the last $k=15$ observations are passed to the judge, and one paragraph
is added to the system prompt stating that text observations carry the same
status as screenshots. Both adapted prompts additionally ask the judge to
report, separately from success, whether the site blocked access (CAPTCHA,
bot detection, region block); this flag does not enter the success label.

\subsection{Search-based evaluation protocol}
\label{app:search-protocol}
Our agents start in search-based mode. Table~\ref{tab:search-retention}
reports results for the base model and \agent on HLE, xbench-DeepSearch-2510,
and BrowseComp-EN, whose full benchmark sets contain $2{,}500$, $100$,
and $1{,}266$ tasks, respectively.

Published baselines in Table~\ref{tab:search-retention} retain their
reported evaluation settings. HLE results reported as text-only are
marked with $^t$; unmarked HLE scores use the full set. QUEST and REDSearcher use their reported context-management results
on BrowseComp-EN. We report xbench-DeepSearch-2510 scores only when that
benchmark version is explicitly identified.

Search-based answers are graded against reference answers. Exact matches
are accepted directly; other answers receive five Kimi-K3 judge votes
at temperature $0.6$, with the majority deciding correctness. Each vote
uses the prompt in Appendix~\ref{app:search-judge-prompt} and
returns a JSON verdict with the extracted final answer and a binary label.

\subsection{Shared settings and benchmark coverage}

\paragraph{Shared execution settings.}
The base model and \agent run in the same \framework Framework with the
same tools, prompts, and sampling parameters: a $100$-step budget per
attempt on operation-based tasks and $300$ steps on search-based tasks,
reasoning enabled, $8{,}192$ output tokens per call, per-call decorrelated
sampling seeds, and the 128K/64K context compression schedule. The
\textsc{switch} action remains available unless explicitly disabled for an
ablation. The \benchmark comparison uses a $100$-step budget for all
systems, as detailed in Appendix~\ref{app:hybrid-protocol}.

\paragraph{Benchmarks.}
Table~\ref{tab:bench} lists the full evaluation sets. Online-Mind2Web and
WebVoyager measure success and efficiency on operation-based tasks;
\benchmark measures hybrid task completion. HLE, xbench-DeepSearch-2510,
and BrowseComp-EN assess retention of deep search capabilities. Every system
in a controlled comparison is evaluated on the same tasks.

\begin{table}[h]
\centering
\small\setlength{\tabcolsep}{4pt}
\begin{tabular}{@{}l r >{\raggedright\arraybackslash}p{6.2cm}@{}}
\toprule
\textbf{Benchmark} & $n$ & \textbf{Protocol} \\
\midrule
Online-Mind2Web \citep{xue2025onlinemind2web}  & 300              & adapted WebJudge; o4-mini \\
WebVoyager \citep{he2024webvoyager}            & 643              & adapted auto-evaluator; GPT-4o \\
\benchmark                                   & 100              & human-written tasks and info/op checklists; at least one op sub-goal per task \\
HLE \citep{phan2025hle}                       & 2{,}500          & reference-answer judge \\
xbench-DeepSearch-2510 \citep{chen2025xbench} & 100              & full 2510 release; reference-answer judge \\
BrowseComp-EN \citep{wei2025browsecomp}       & 1{,}266          & reference-answer judge \\
\bottomrule
\end{tabular}
\caption{Benchmark sizes and scoring protocols. $n$ is the full benchmark
size.}
\label{tab:bench}
\end{table}

\section{Switching patterns and example tasks}
\label{app:archetypes}

Table~\ref{tab:archetypes} summarizes the four patterns and their
difficulty profiles; one naturalized task from the design run is shown for
each. Difficulty is realized by the number of bound sites and the number of
cross-site \emph{bridge} rounds (a bridge is an $\modeS$ prototype that
consumes an $\modeO$ output from one site and produces the input for an
$\modeO$ prototype on another).
Medium and hard $\mathrm{A}_3/\mathrm{A}_4$ tasks chain phases across
sites with one or two bridge rounds, yielding two to five executed switches.

\begin{table}[h]
\centering
\small
\setlength{\tabcolsep}{4pt}
\begin{tabular}{@{}l l >{\raggedright\arraybackslash}p{1.6cm} >{\raggedright\arraybackslash}p{2.7cm} >{\raggedright\arraybackslash}p{6.0cm}@{}}
\toprule
\textbf{Pattern} & \textbf{Start} & \textbf{Working modes} & \textbf{Difficulty} & \textbf{Switching motivation} \\
\midrule
$\mathrm{A}_1$ & $\modeO$ & $\modeS$ & standard (1 site) & The opening decision corrects from $\modeO$ to $\modeS$ before any browser work: the task is plainly search-answerable. \\
$\mathrm{A}_2$ & $\modeS$ & $\modeS,\modeO$ & standard (1 site) & A bounded, good-faith $\modeS$ attempt (two materially different exact queries plus one result inspection), then fall back to $\modeO$ once structural insufficiency is established. \\
$\mathrm{A}_3$ & $\modeS$ & $\modeS,\modeO$ & easy / medium / hard (1 / 2 / 3 sites) & $\modeS$ produces an entity, constraint, identifier, threshold, or shortlist that $\modeO$ requires. \\
$\mathrm{A}_4$ & $\modeO$ & $\modeO,\modeS$ & easy / medium / hard (1 / 2 / 3 sites) & $\modeO$ surfaces an entity, offer, item, metric, or status that binds the subsequent $\modeS$ question. \\
\bottomrule
\end{tabular}
\caption{Switching patterns and difficulty levels. $\modeS$/$\modeO$
denote search-based/operation-based modes. Start is the initial mode;
Working modes lists the modes used to perform the task.}
\label{tab:archetypes}
\end{table}

\paragraph{$\mathrm{A}_1$ (standard).}
\task{I'm looking at three stocks for a possible portfolio addition:
SoundHound AI (SOUN), BigBear.ai (BBAI), and C3.ai (AI). I want to compare
their current market data side by side---current price, market cap, P/E
ratio, 52-week high and low, and average volume---so I can evaluate which
one looks most promising.} The task starts on a finance site, but every
requested value is published on indexed pages; the intended decision is to
switch to $\modeS$ immediately.

\paragraph{$\mathrm{A}_2$ (standard).}
\task{I need to check the voting status of Uniswap governance proposals that
are still active. Find any proposals whose voting deadline falls on or after
August 15, and report the proposal title, status, current For and Against
vote counts, deadline, and quorum requirement.} Search surfaces proposal
titles and discussion threads but not live vote tallies; after a bounded
attempt the intended decision is to switch to the governance dashboard.

\paragraph{$\mathrm{A}_3$ (medium, two sites).}
\task{I'm planning a trip in October and need to get from a small town near
Lyon to the nearest major TGV hub. First, find which major TGV station is
closest to Vienne, France. Then use Wanderlog to map the driving route from
Vienne to that station. After that, look up the transit fare policy from
that station to Geneva, and finally use Transit App to check the actual
public-transit route and fare.} $\modeS$ resolves the station and the fare
policy; each $\modeO$ phase consumes an $\modeS$ output.

\paragraph{$\mathrm{A}_4$ (medium, two sites).}
\task{I'm a senior machine-learning engineer looking at both Apple and
Airbnb. Find the senior ML engineer opening at Apple that lists PyTorch and
TensorFlow as required skills; once you have that role and its job ID, look
up how Airbnb's career site labels the equivalent role family so you can
search for the matching opening on Airbnb's careers page and compare the two
listings.} $\modeO$ surfaces the Apple listing; $\modeS$ bridges the role
family; $\modeO$ consumes it on the second site.

\section{Trajectory review and refinement}
\label{app:judges}

The reviewers and refiners in the multi-agent synthesis framework verify and refine
the executor's trajectories. Table~\ref{tab:judges} lists their
checks, inputs, and outputs.
Every quotation a judge makes is mechanically verified against the
trajectory; a quotation that cannot be located voids the judgement.

\begin{table}[h]
\centering
\small
\setlength{\tabcolsep}{4pt}
\begin{tabular}{@{}>{\raggedright\arraybackslash}p{3.0cm} >{\raggedright\arraybackslash}p{4.4cm} >{\raggedright\arraybackslash}p{5.6cm}@{}}
\toprule
\textbf{Judgement} & \textbf{Input} & \textbf{Consumption} \\
\midrule
Unified audit (two stages) & rollout, contract, final answer; screenshots for $\modeO$ phases & per-phase progress account; leakage steps $\to$ rewrite; task-success and route verdicts $\to$ filter \\
Leakage rewrite & leakage steps and their reasoning & reasoning replaced in place under fact/decision/coherence constraints, then re-verified \\
Residue detection & controller's feedback log (which message reached which step) & second rewrite pass; trajectories with pervasive leakage dropped \\
Site fidelity ($\mathrm{A}_1$) & final values, observations & drop if the answer was not obtained from the named site (judged value by value) \\
Switch rationale ($\mathrm{A}_1/\mathrm{A}_2$) & rewritten reasoning at the switch step & drop if the switch was not triggered by the pattern's intended condition \\
Answer coverage & answer schema, all observations & drop if a required field is unobtainable; otherwise merge fields obtained in earlier phases (numeric values re-verified against observations) \\
$\modeO$-segment purity & $\modeO$ segments with screenshots & drop the segment if no browser-only interaction was performed \\
\bottomrule
\end{tabular}
\caption{Trajectory review checks.}
\label{tab:judges}
\end{table}

\clearpage

\section{Pipeline algorithms}
\label{app:algorithms}

Algorithm~\ref{alg:pipeline} summarizes the synthesis pipeline of
\Secref{sec:method-data} end to end, and Algorithm~\ref{alg:rollout} expands
the collaboration between the executor and controller in Stage~3.
This stage also includes task synthesis and trajectory review, as shown in
Algorithm~\ref{alg:pipeline}.
Data collection models and task validation checks are described in
Appendix~\ref{app:collection-details}.

\begin{algorithm}[t]
\caption{\datapipe trajectory synthesis pipeline}
\label{alg:pipeline}
\begin{algorithmic}[1]
\Require site set $W$; pattern and difficulty ratios; probe budget $B$; task synthesizer; executor $\pi_E$; controller $\nu$; reviewers and refiners
\State \textbf{Stage 1 --- task prototypes from capability probing}
\ForAll{sites $w\in W$}
  \State $F_w\gets\mathrm{Explore}(w)$ \Comment{operator drives the browser; extractor records answer-shaped facts}
\EndFor
\State $F\gets\mathrm{ValueFilter}\big(\bigcup_w F_w\big)$ \Comment{keep concrete values; drop scaffolding and duplicates}
\ForAll{facts $f\in F$}
  \State $\ell(f)\gets\mathbf{1}[\mathrm{Probe}_{\modeS}(f;B)\ \text{recovers}\ f]$ \Comment{Eq.~(\ref{eq:probe}); transcript retained}
  \State label $f$ as search-reachable / intrinsically operation-bound / prober-limited / volatile
\EndFor
\State remove unstable or unreliable facts; form $F_{\modeS}$ and $F_{\modeO}$
\State abstract prototype vocabularies $P_{\modeS},P_{\modeO}$; retain website and entity bindings
\State \textbf{Stage 2 --- switching pattern design}
\State define $\Phi=\{\mathrm{A}_1,\mathrm{A}_2,\mathrm{A}_3,\mathrm{A}_4\}$ \Comment{efficiency, search failure, and information transfer in both directions}
\State \textbf{Stage 3 --- multi-agent trajectory synthesis}
\ForAll{blueprints to generate}
  \State sample pattern $\phi$, sites $w$, entity $e$, difficulty $d$; sample composition $c$ over admissible pairs $\mathrm{out}(p_{\modeS})\in\mathrm{in}(p_{\modeO})$ \Comment{Eq.~(\ref{eq:contract}); deterministic given the seed}
  \State build contract $C(b)=((g_k,\delta_k,m_k))_{k=1}^K$ with produce/consume edges \Comment{Eq.~(\ref{eq:contract})}
  \Repeat
    \State $x\gets\mathrm{Naturalize}(b)$ \Comment{question, per-phase goals, switch triggers, answer schema}
  \Until{$\mathrm{Validate}(x,b)$ passes or the retry limit is reached}
\EndFor
\ForAll{tasks $(x,C(b))$}
  \State $\tau\gets\textsc{CoordinatedExecution}(x,C(b))$ \Comment{Algorithm~\ref{alg:rollout}}
  \State \textbf{if} $\tau$ did not end in \textsc{answer} \textbf{then continue}
  \State $(\text{progress},L,\text{verdicts})\gets\mathrm{Audit}(\tau,C(b))$ \Comment{two stages; every quotation verified verbatim}
  \State \textbf{if} task not successful \textbf{then} discard $\tau$
  \ForAll{leakage steps $i\in L$} \State $r_i\gets\mathrm{Rewrite}(r_i)$; re-verify \Comment{facts and decision unchanged}
  \EndFor
  \State detect residual leakage from the controller's feedback log; rewrite again or discard if pervasive
  \State drop rejected intermediate \textsc{answer} steps
  \State \textbf{if} site fidelity, switch rationale, or answer coverage fails \textbf{then} discard $\tau$ (or merge fields obtained in earlier phases)
  \ForAll{$\modeO$ segments $\varsigma$ of $\tau$} \State \textbf{if} $\varsigma$ performed no browser-only interaction \textbf{then} drop $\varsigma$
  \EndFor
  \State $\mathcal{D}\gets\mathcal{D}\cup\mathrm{Export}(\tau)$ \Comment{Eq.~(\ref{eq:export}): $n_{\mathrm{sw}}+1$ samples}
\EndFor
\State \Return $\mathcal{D}$
\end{algorithmic}
\end{algorithm}

\begin{algorithm}[t]
\caption{\textsc{CoordinatedExecution}: collaboration between the executor and controller}
\label{alg:rollout}
\begin{algorithmic}[1]
\Require task $x$; contract $C=((g_k,\delta_k,m_k))_{k=1}^K$; executor $\pi_E$; controller $\nu$; sanitizer $\sigma$; per-phase action budgets; step budget $T_{\max}$
\State $k\gets1$;\ \ $m\gets m_1$;\ \ $\beta\gets\emptyset$;\ \ $u\gets\emptyset$ \Comment{phase, mode, verified bindings, last control-agent message}
\For{$t=1,\dots,T_{\max}$}
  \State $a_t\sim\pi_E\big(\cdot\mid x,g_k,\beta,\sigma(u),\tau_{<t}\big)$ rendered through $\mathcal{R}$ \Comment{Eq.~(\ref{eq:execution-control}); plan details hidden}
  \If{$a_t=\textsc{answer}(y)$}
     \State \textbf{if} $k=K$ and $\nu(\delta_K,\tau_K)=1$ \textbf{then} \Return $\tau$ \textbf{else} $u\gets$ rejection (answer step recorded, later dropped); \textbf{continue}
  \ElsIf{$a_t=\textsc{switch}$}
     \If{$\nu(\delta_k,\tau_k)=1$}
        \State $\beta\gets\beta\cup\mathrm{bindings}(\tau_k)$; pin cited evidence steps against folding; $k\gets k{+}1$; $m\gets\bar m$; open a new segment
     \Else
        \State $u\gets$ rejection; record the step with \texttt{switch\_executed}=false \Comment{no new segment}
     \EndIf
  \Else
     \State $o_t\gets\mathrm{Env}_m(a_t)$; append $(m,a_t,o_t)$
     \State \textbf{if} a human-verification gate is detected \textbf{then} terminate
     \If{phase budget exhausted, or no progress for three actions, or eight actions since the last check}
        \State $u\gets\sigma\big(\nu\text{'s assessment of }(\delta_k,\tau_k)\big)$ \Comment{contract vocabulary stripped before it reaches $\pi_E$}
     \EndIf
  \EndIf
\EndFor
\State \Return $\tau$ \Comment{terminated by budget}
\end{algorithmic}
\end{algorithm}

\clearpage

\section{Training details}
\label{app:training}

Both data sources are rendered into multi-turn SFT samples using the same
code paths as inference. System prompts, history blocks, context compression,
and screenshot placement follow \Secref{sec:method-harness}.
Samples that exceed the context budget even after folding are split into
two or three parts by greedy bisection rather than dropped, preserving
long-horizon supervision. Table~\ref{tab:training} gives the training settings.

\begin{table}[H]
\centering
\small\setlength{\tabcolsep}{4pt}
\begin{tabular}{@{}>{\raggedright\arraybackslash}p{4.2cm} >{\raggedright\arraybackslash}p{9.2cm}@{}}
\toprule
\textbf{Setting} & \textbf{Value} \\
\midrule
Backbone & Qwen3.6-35B-A3B \citep{qwen36_35b_a3b} \\
Trainable parameters & language model; vision encoder and visual projector frozen \\
Objective & next-token cross-entropy \\
Sample unit & one segment (context rendered by the harness; targets = the segment's assistant turns) \\
Context length & 131{,}072 tokens (folded to the 64K watermark when building samples) \\
Reasoning tokens & retained and supervised \\
Data mixture & hybrid samples and $7{,}000$ public search-based trajectories, stored in one shuffled dataset \\
Optimizer & Adam; $\beta_1=0.9$, $\beta_2=0.95$, $\epsilon=10^{-8}$ \\
Learning rate & peak $2\times10^{-5}$; minimum $2\times10^{-6}$ \\
Schedule & $5\%$ warmup from zero, followed by cosine decay \\
Weight decay & $0.1$ (constant) \\
Gradient clipping & $1.0$ \\
Training epochs & $1$ \\
Effective batch size & $32$ samples per parameter update \\
Precision & BF16 \\
Packing / padding-free mode & disabled / enabled \\
Random / data seed & $42$ / $42$ \\
Data shuffling / validation split & enabled / $0$ \\
Activation recomputation & full, uniform, one layer per recomputation group \\
Training framework & Megatron via ms-swift \texttt{4.5.0.dev0} \\
\bottomrule
\end{tabular}
\caption{Training configuration.}
\label{tab:training}
\end{table}

\section{Baseline details}
\label{app:baselines}
We describe the baselines in Tables~\ref{tab:main}, \ref{tab:hwb-single},
\ref{tab:hybrid}, and \ref{tab:search-retention}. Published benchmark scores
serve as external references; our evaluations on \benchmark compare agents
on the same tasks and completion checklists. Evaluation protocols and
differences in benchmark coverage are given in Appendix~\ref{app:eval};
base-model and framework controls are in Appendix~\ref{app:controls}.

\subsection{Open-source operation-based agents}

These agents choose browser actions from visual observations. All four
families appear in both the traditional operation-based evaluation and
the comparison on \benchmark; the GUI-Owl model size differs between
these settings.

\paragraph{Fara-7B.}
Fara-7B \citep{awadallah2025fara} is a 7B computer-use agent that reads
screenshots and predicts mouse and keyboard actions. Its FaraGen data
pipeline generates tasks on real websites, collects multiple execution
attempts, and filters trajectories using several verifiers. It provides
a compact operation-based baseline trained on synthetic web interactions.
We use the Fara-7B release.

\paragraph{UI-TARS-1.5-7B.}
UI-TARS-1.5 \citep{qin2025uitars,bytedance2025uitars15} integrates visual
perception, reasoning, and action prediction in a vision-language model.
It reasons before acting and uses reinforcement learning to improve GUI
interaction. Actions include coordinate-based clicks, typing, and scrolling.
Our baseline is the publicly released 7B model, which tests whether a
general GUI agent can solve the web tasks through browser interaction.

\paragraph{GUI-Owl-1.5.}
GUI-Owl-1.5 \citep{xu2026mobileagentv35} supports browser, desktop, and
mobile interaction. Its training combines synthetic interaction data
with reinforcement learning across environments, covering visual
grounding, planning, and action execution. We include
GUI-Owl-1.5-8B-Thinking on Online-Mind2Web and WebVoyager
(Table~\ref{tab:main}), and the 32B variant on WebVoyager
(Table~\ref{tab:main}) and \benchmark (Table~\ref{tab:hwb-single}).
The 32B variant has no published Online-Mind2Web result, so that
benchmark is omitted for it in Table~\ref{tab:main}.

\paragraph{MolmoWeb-8B.}
MolmoWeb \citep{gupta2026molmoweb} predicts browser actions from task
instructions and screenshots. Its MolmoWebMix training data includes
synthetic trajectories, human demonstrations, and examples of individual
web skills and visual grounding. It interacts through the rendered page
without requiring HTML or an accessibility tree. We use the 8B model
as an open visual web-agent baseline.

\subsection{Open-source search-based agents}

All eight agents below appear in the deep search comparison
(Table~\ref{tab:search-retention}). We additionally evaluate
XYZ-Aquila-mini and Agents-A1 with search-based tools on \benchmark
(Table~\ref{tab:hwb-single}). This classification describes their role
in our evaluation; some models also support other agentic tasks.

\paragraph{Tongyi DeepResearch.}
Tongyi DeepResearch \citep{tongyi2025deepresearch} is a 30B-class
mixture-of-experts research agent with roughly 3B active parameters.
It learns long search-and-reading trajectories through agentic
mid-training, supervised fine-tuning, and reinforcement learning.
Its training uses automatically generated tasks and trajectories, while
context management supports extended information gathering. We include
the published scores of its standard open-model setting.

\paragraph{OpenSeeker-v1.}
OpenSeeker \citep{du2026openseeker} builds on
Qwen3-30B-A3B-Thinking-2507 and uses supervised fine-tuning on about
11.7K synthetic trajectories. Its task generation links facts across web
pages and hides identifying clues to require multi-step search. This
provides a search-based baseline trained with an open data pipeline and
SFT alone. We report the first release, OpenSeeker-v1.

\paragraph{REDSearcher.}
REDSearcher \citep{chu2026redsearcher} constructs difficult search tasks
by controlling how evidence is connected and distributed across sources.
It combines mid-training for evidence grounding, planning, and tool use
with supervised fine-tuning and reinforcement learning. A simulated
search environment reduces the cost of training rollouts. We include
the published 30B search-agent results.

\paragraph{MiroThinker-1.7-mini.}
MiroThinker-1.7 \citep{miromind2026mirothinker17} targets long research
tasks with agentic mid-training focused on planning, reasoning over
accumulated context, and tool interaction. These objectives help the
agent sustain information gathering across many steps. We include
the open 30B MiroThinker-1.7-mini model.

\paragraph{FORT-Searcher.}
FORT-Searcher \citep{deng2026fortsearcher} focuses on training tasks
that cannot be solved through easy shortcuts. Its data pipeline controls
entity selection, evidence connections, and question wording, then
refines questions to remove alternative routes that bypass the intended
search. It trains on the resulting trajectories using supervised
fine-tuning alone. Our comparison uses the published 30B model.

\paragraph{QUEST-35B.}
QUEST \citep{xie2026quest} trains deep research agents with fully
synthetic tasks. It organizes evaluation criteria into rubric trees
to provide rewards for both factual questions and open-ended research,
and combines mid-training, supervised fine-tuning, and reinforcement
learning. Its tasks cover evidence gathering and report writing.
We include the 35B model's published search benchmark results.

\paragraph{Agents-A1.}
Agents-A1 \citep{bai2026agentsa1} is a 35B mixture-of-experts model
trained for long-horizon tasks across several agentic domains. Its
post-training combines broad supervised data with domain-specific
expert knowledge and long interaction trajectories. We include its
published deep search results and evaluate its search-based
configuration on \benchmark, providing a baseline with the same
total parameter scale as \agent.

\paragraph{XYZ-Aquila-mini.}
XYZ-Aquila-mini \citep{xyz2026ai4ai} builds on Qwen3.6-35B-A3B and
targets deep search through improvements to training data and the
execution framework. Its pipeline generates tasks from connected web
evidence, preserves tool observations in supervised trajectories, and
refines context management using verified evaluation feedback. We
include its published deep search scores and evaluate its search-based
configuration on \benchmark. It shares the backbone family of \agent.

\subsection{Frontier LLM systems}

\paragraph{Operation-based references.}
Table~\ref{tab:main} includes published results for GPT-5.5,
GPT-5 (SoM), o3 (SoM), Claude-Opus-4.6, Seed2.1-Pro, GLM-5V-Turbo,
Kimi-K2.6, and Qwen3.8-Max. These rows represent models paired with
the browser interfaces used in their respective evaluations. SoM
denotes a set-of-marks interface that labels actionable screen elements.
We also include Gemini 2.5 Computer Use, OpenAI Operator, and
Yutori Navigator (n1), which are systems designed for computer or
browser interaction. These results provide frontier references under
their reported settings.

\paragraph{Deep search references.}
Table~\ref{tab:search-retention} includes GPT-5.6 Sol, Claude Fable 5,
Qwen3.8-Max, Kimi-K3, Kimi-K2.6, GLM-5.2, and
DeepSeek-V4-Pro (Preview). Their published scores describe performance
with tool-assisted information gathering and reasoning. These rows
provide context for the open-source search agents; differences in
evaluation sets are retained in the table annotations and described
in Appendix~\ref{app:eval}.

\paragraph{Frontier models within \framework.}
We evaluate Kimi-K3, Qwen3.8-Max, and Qwen3.7-Plus zero-shot within
\framework on \benchmark. Each has access to both search-based and
operation-based modes through the framework. This comparison tests
how much hybrid task performance can be obtained from frontier models
without our hybrid training. These are our evaluations, separate from
the published reference rows above.

\section{Framework implementation details}
\label{app:harness}

\subsection{Tool functions and signatures}
Table~\ref{tab:tool-sig} lists tool functions and arguments, grouped into
search-based tools, operation-based tools, and tools shared by both modes.
Search-based APIs return text; browser
actions return a screenshot and status after execution. The reason for
\texttt{switch} appears in the preceding reasoning.
Coordinates use a $0$--$1000$ logical grid over a $1280\times720$ viewport
and are converted to pixels by the runtime. This convention is learned
from data and is not described in the prompt.

The \texttt{search} and \texttt{scholar} tools use the Serper API.
The \texttt{visit} tool uses the Jina API to fetch page content and
\texttt{deepseek-flash-0731} to summarize it according to the extraction goal.

\begin{table}[!htb]
\centering
\footnotesize
\setlength{\tabcolsep}{4pt}
\begin{tabular}{@{}>{\raggedright\arraybackslash}p{3.1cm} >{\raggedright\arraybackslash}p{4.0cm} >{\raggedright\arraybackslash}p{6.2cm}@{}}
\toprule
\textbf{Tool} & \textbf{Arguments} & \textbf{Function} \\
\midrule
\rowcolor{groupbg}\multicolumn{3}{l}{\textbf{Search-based tools ($\modeS$)}} \\
\texttt{search} & \textbf{query} (string or list) & Search the web; return ranked snippets with titles, URLs, sources, and dates. \\
\texttt{scholar} & \textbf{query} (string or list) & Search academic sources; return publication details, citation counts, and PDF links. \\
\texttt{visit} & url (string or list), \textbf{goal}, parse\_type & Extract goal-relevant text from pages or PDFs; defaults to the current URL and does not change browser state. \\
\midrule
\rowcolor{groupbg}\multicolumn{3}{l}{\textbf{Operation-based tools ($\modeO$)}} \\
\texttt{goto} / \texttt{new\_tab} & \textbf{url} & Open a URL in the current or a new browser tab. \\
\texttt{go\_back} & none & Return to the previous page in the current tab. \\
\texttt{switch\_tab} / \texttt{close\_tab} & \textbf{index} & Select or close the specified browser tab. \\
\texttt{click} / \texttt{double\_click} / \texttt{hover} & \textbf{x}, \textbf{y} & Click, double-click, or hover at the specified coordinates. \\
\texttt{type} & \textbf{text}, submit & Enter text, optionally submitting it. \\
\texttt{press\_key} & \textbf{keys} & Press a key or key combination. \\
\texttt{select\_option} & \textbf{x}, \textbf{y}, \textbf{value} & Select a value in a dropdown at the specified coordinates. \\
\texttt{scroll} & \textbf{direction}, x, y & Scroll at a position, or scroll the page if no position is supplied. \\
\texttt{drag} & \textbf{from\_x}, \textbf{from\_y}, \textbf{to\_x}, \textbf{to\_y}, hold\_ms & Drag between two positions, optionally holding before release. \\
\texttt{wait} & time & Wait for the specified duration. \\
\midrule
\rowcolor{groupbg}\multicolumn{3}{l}{\textbf{Shared tools ($\modeS$ and $\modeO$)}} \\
\texttt{code\_sandbox} & \textbf{code}, language, stdin, run\_timeout, compile\_timeout, files, fetch\_files & Run isolated code for calculation, parsing, and validation. \\
\texttt{switch} & none & Change mode, replace its tool interface, and rebuild context; return a hand-off notice without executing an environment action. \\
\bottomrule
\end{tabular}
\caption{\framework tool functions and arguments, grouped by mode.
Bold arguments are required; all other arguments are optional.}
\label{tab:tool-sig}
\end{table}

\subsection{Runtime parameters}
\label{app:runtime}
Table~\ref{tab:params} lists the runtime parameters shared by the base
and fine-tuned rows of Table~\ref{tab:main}.

\begin{table}[h]
\centering
\small
\begin{tabular}{@{}l l@{}}
\toprule
\textbf{Parameter} & \textbf{Value} \\
\midrule
Viewport & $1280\times720$ \\
Coordinate grid & $0$--$1000$ (converted to pixels by the runtime) \\
Step budget $T_{\max}$ per attempt (operation-based) & 100 \\
Max output tokens per turn & 8{,}192 \\
Temperature / top-$p$ / top-$k$ & 1.0 / 0.95 / 20 \\
Repetition penalty & 1.05 \\
Fold watermarks $\theta_{\mathrm{hi}}/\theta_{\mathrm{lo}}$ & 131{,}072 / 65{,}536 tokens \\
Folded-observation prefix $c$ & 600 characters \\
Screenshots retained in context & all unfolded $\modeO$ steps \\
Search results per query & top 10 \\
\texttt{visit} extraction & goal-conditioned summary, up to 32{,}000 tokens per chunk \\
Format retries per turn & 5 (fresh seed each) \\
Error/format-failure attempts & up to 2 \\
Attempts ending at the step limit & up to 5, restarting with a fresh context \\
Per-call timeouts (agent / summarizer) & 1{,}800\,s / 1{,}800\,s \\
Seed derivation & base seed $+$ crc32(task, attempt) $+ 7919\cdot t$ \\
\bottomrule
\end{tabular}
\caption{\framework runtime parameters for the base model and \agent
in Table~\ref{tab:main}.}
\label{tab:params}
\end{table}

\subsection{Context compression}
\label{app:compression}
Compression is triggered by two watermarks on the token count
$T(\mathcal{R})$ of the rendered context (text tokens, image tokens, and
chat-template overhead):
\begin{equation}
\text{if } T(\mathcal{R})>\theta_{\mathrm{hi}}\ \text{ then fold until } T(\mathcal{R})\le\theta_{\mathrm{lo}},
\qquad
\theta_{\mathrm{hi}}=131{,}072,\ \ \theta_{\mathrm{lo}}=65{,}536,
\label{eq:fold}
\end{equation}
where ``fold'' means truncating a step's observation to its first
$c=600$ characters and dropping its reasoning and screenshot. The trigger actually
used reserves room for generation,
$\theta_{\mathrm{hi}}^{\mathrm{eff}}=\min(\theta_{\mathrm{hi}},L_{\max})-(n_{\mathrm{out}}+256)$,
with $L_{\max}$ the serving window and $n_{\mathrm{out}}$ the maximum output
length. Folding proceeds in a fixed order: first the steps already in the
history block, oldest first; only when all of them are folded is the segment
start $s(t)$ advanced, which moves the oldest steps of the current segment
into the history block, entering it already folded. The most recent step is
never folded, tool calls are never altered, and steps cited as evidence by
the controller during data collection are exempt. The gap between
$\theta_{\mathrm{hi}}$ and $\theta_{\mathrm{lo}}$ exists for the key--value
cache: a fold rewrites the prefix, so folding ``just enough'' at every step
would invalidate the cache at every step, whereas folding down to
$\theta_{\mathrm{lo}}$ once buys a long run of pure appends. Screenshots
follow the same rule---kept for unfolded steps, dropped once a step is
folded---and observations in the search-based mode never carry images.

\section{Data collection implementation details}
\label{app:collection-details}

We use Qwen3.8-Max and Kimi-K3 for the LLM-based components of
\datapipe.

\subsection{Naturalization validators}
\label{app:validators}

A naturalized task is accepted only if all of the following hold; a failure
returns the reason to the generator, which re-samples.
\begin{enumerate}\setlength{\itemsep}{1pt}
\item The question is non-trivial in length and contains no meta-language
about agent routing or mode switching.
\item The site bindings in the generated text preserve the blueprint's fixed
site domains exactly.
\item The per-component goals and completion conditions cover the
blueprint's components exactly, with no empty entries.
\item No component's goal or completion condition names a site outside its
declared site scope, except the site of a producer it directly consumes
from.
\item For medium and hard $\mathrm{A}_3/\mathrm{A}_4$ tasks, the question
explicitly names every target site.
\item The switch triggers cover every switch policy in the blueprint, with no
empty entries.
\item A coherence rationale and a non-empty answer schema are present.
\item The dependency bindings match the blueprint's dependency edges exactly.
\item Producer answers remain unknown in the question text until their phase
completes (no leakage of downstream values).
\end{enumerate}

\clearpage

\section{System and evaluation prompts}
\label{app:prompt-listings}

\subsection{Mode-specific system prompts}
\label{app:prompts}
The two system prompts share the output-format block; the mode-specific
parts are reproduced below verbatim. The answer-shape sentence differs by
mode: in $\modeS$ it asks for a concise entity, number, or list; in $\modeO$
it additionally allows stating the end state reached when the task was an
operation. Tool schemas (Table~\ref{tab:tool-sig}) are appended after the
prompt.

\paragraph{Search-based mode ($\modeS$).}
\begin{scriptsize}
\begin{verbatim}
You are a general-purpose agent on the SEARCH-BASED path.

Your tools this turn: search, scholar, visit, code_sandbox, switch.
- `search` finds sources and verifies facts; it returns ranked results with snippets.
- `scholar` is for academic papers, citation-rich evidence and PDFs.
- `visit` opens a specific URL and extracts the facts you ask for via `goal`.
- `code_sandbox` is for calculation, parsing and validation - never for browsing.
- `switch` hands control to the operation-based (browser) path. Use it only when
  the needed information requires real page interaction (login, form, dynamic
  content that search and visit cannot reach). It takes no arguments - the reason
  belongs in `<think>`, do not restate it as a parameter.

# Avoiding Dead Ends (read this before every action)

Your earlier steps are recorded in the task message below - every query you ran
is listed there, though older observations may be truncated to save space.
Use it. Before acting, check that history:

1. NEVER re-run a query that is a near-duplicate of one already in the history.
   If a formulation failed, change the approach, not the wording: switch
   language, search the concrete entity/attribute instead of the task's
   paraphrase, add a site: filter or an authoritative source, use quoted exact
   phrases, boolean OR over alternatives, or negative terms to exclude the
   wrong candidates.

2. NEVER stay on one candidate for more than ~5 consecutive actions. If your
   last several actions all revolve around the same entity and none of them
   added a NEW confirmed fact, you are stuck. On the next turn you MUST do one
   of these instead:
   - name at least two ALTERNATIVE candidates that also satisfy the constraints
     you have already confirmed, and search for one of them;
   - run a DISCONFIRMING search for your current candidate - actively look for
     evidence that it VIOLATES one of the task's constraints;
   - go back to the task statement, list the constraints you have NOT verified
     yet, and search for one of those directly.

3. Search snippets are LEADS, not evidence. If a snippet looks like it contains
   the answer or a decisive fact, open that page with `visit` before you rely on
   it. Do not build a final answer out of snippets alone.

4. Work the constraints in parallel, not strictly in order. A multi-constraint
   task is usually cracked by finding the ONE constraint that is rare enough to
   pin down a small candidate set - start from the most distinctive constraint,
   and if a line stalls, move to another constraint rather than pushing harder
   on the same one.

5. If a page fails to load or returns nothing usable, do not retry the same URL.
   Find another source for the same fact.
\end{verbatim}
\end{scriptsize}

\paragraph{Operation-based mode ($\modeO$).}
\begin{scriptsize}
\begin{verbatim}
You are a general-purpose agent on the OPERATION-BASED path (real browser).

You act on the live page: goto, new_tab, click, double_click, hover, type,
select_option, scroll, drag, press_key, wait, go_back, switch_tab, close_tab.
`code_sandbox` is available for calculation/parsing. `switch` hands control back
to the search-based path; it takes no arguments - put the reason in `<think>`.

Once a page is open you get its screenshot each turn - ground every action in
what is actually visible there. If no page is open yet, `goto` a URL first;
there is nothing on screen to click.

# Avoiding Dead Ends (read this before every action)

Your earlier actions and the resulting pages are recorded in the task message
below (older observations may be truncated). Use it.

1. NEVER click the same coordinate twice. If a click produced no visible change
   (same page, same layout), that element is not what you thought it was - do
   NOT click it again, and do NOT click a point a few pixels away hoping for a
   different result. Re-read the current screenshot and pick a different target,
   or reach the goal another way (scroll to reveal more, use the page's own
   navigation/search box, or `goto` a direct URL).

2. If two consecutive actions produced no state change, stop repeating that
   interaction pattern. Change the approach: scroll, open a different section,
   use the site's search, or `go_back` and take another route.

3. Before typing, make sure the target field is actually focused - click it
   first, verify from the screenshot that the caret/highlight is in the field,
   then type.

4. If the current O-side page clearly presents a CAPTCHA, Cloudflare/Turnstile
   challenge, "verify you are human" gate, or any other human-verification step,
   STOP immediately. Do not click, wait, refresh, retry, work around it, or
   `switch` to another path. End the entire task with `<answer>`, briefly stating
   that human verification blocked the requested operation and what progress
   was completed before the block. Human verification is a terminal condition
   even when the requested result is still incomplete.

5. If the page needs information you cannot reach by interaction for another
   reason (paywall, login wall, dead link, infinite scroll with no matches), do
   not keep clicking - `switch` back to the search-based path and find the fact
   elsewhere.

6. Verify state changes from the screenshot before assuming success. Do not
   chain several blind actions on the assumption that an earlier one worked.
\end{verbatim}
\end{scriptsize}

\paragraph{Shared output format (both modes).}
\begin{scriptsize}
\begin{verbatim}
# Output Format

Every turn, output exactly:

<think>
Your reasoning for this turn: what the last observation told you, what is still
unverified, and why this specific next action.
</think>
<tool_call>
{"name": "<tool name>", "arguments": {...}}
</tool_call>

Call at most ONE tool per turn. The next message contains that tool's result.
[...]
When you have enough verified evidence, stop calling tools and output:

<think>
Why the evidence is now sufficient - map each constraint in the task to the fact
that confirms it.
</think>
<answer>
your final answer
</answer>

# Tool call OR answer - never both

Each turn ends in EXACTLY ONE of two ways: a `<tool_call>`, or an `<answer>`.
Emitting both in the same turn is an error. [...]
\end{verbatim}
\end{scriptsize}

\subsection{Search-based answer judging}
\label{app:search-judge-prompt}
The reference-answer judge in Appendix~\ref{app:search-protocol} uses
the following prompt for each vote.
\begin{scriptsize}
\begin{verbatim}
You are a strict benchmark evaluator.

Your task is to judge whether an agent's final answer matches the provided
reference answer for the given question.

Rules:
1. First extract the agent's final committed answer from the agent output.
2. Judge only whether that extracted final answer matches the reference answer.
3. Ignore extra explanation, evidence, or reasoning unless it changes or
   contradicts the final answer.
4. Accept semantically equivalent answers even if they differ in formatting,
   capitalization, punctuation, articles, common abbreviations, or obvious
   unit/currency/date formatting.
5. For numbers, accept equivalent numeric values after straightforward
   normalization, but reject materially different values, incompatible units,
   or unsupported rounding.
6. For list or set answers, all required items must be present and no
   materially incorrect extra items should be introduced.
7. If the agent gives multiple conflicting answers, or never clearly commits
   to a final answer, mark it incorrect.
8. If there is no answer, set extracted_final_answer to null and
   predicted_label to 0.

Return JSON only, with exactly these keys:
{
  "extracted_final_answer": "string or null",
  "reasoning": "brief explanation",
  "predicted_label": 0
}
\end{verbatim}
\end{scriptsize}

\clearpage

\section{Full task prototype vocabularies}
\label{app:prototypes}

Stage~1 of the pipeline (\Secref{sec:method-data}) abstracts two task
prototype vocabularies from the probed fact pool; the task synthesizer in
Stage~3 composes them into tasks. Each operation
prototype names a recurring kind of interaction that produces
operation-bound facts and declares a typed interface; each search prototype
names a recurring kind of information need satisfiable from the indexed web.
The interface types are: for $\modeS$ outputs, \texttt{entity},
\texttt{candidate\_set}, \texttt{location}, \texttt{date\_window},
\texttt{constraint\_set}, \texttt{identifier}, \texttt{threshold},
\texttt{classification}, and \texttt{fact\_set}; for $\modeO$ outputs,
\texttt{live\_offer}, \texttt{quote}, \texttt{ranked\_item},
\texttt{configured\_item}, \texttt{computed\_result}, \texttt{route},
\texttt{itinerary}, \texttt{cart\_total}, \texttt{metric},
\texttt{live\_status}, \texttt{protocol\_quote}, \texttt{template},
\texttt{visual\_item}, \texttt{record}, \texttt{service\_quote},
\texttt{build}, \texttt{plan}, and \texttt{artifact}.
Tables~\ref{tab:proto-o} and~\ref{tab:proto-s} list the full vocabularies.

A pair
$(p_{\modeS},p_{\modeO})$ is admissible when
$\mathrm{out}(p_{\modeS})\in\mathrm{in}(p_{\modeO})$.

{\footnotesize\setlength{\tabcolsep}{3pt}
\begin{longtable}{@{}>{\raggedright\arraybackslash}p{2.6cm} >{\raggedright\arraybackslash}p{4.5cm} >{\raggedright\arraybackslash}p{2.8cm} >{\raggedright\arraybackslash}p{3.3cm}@{}}
\caption{Operation-based task prototypes $P_{\modeO}$ ($48$ total).
\emph{out}: the prototype's output type; \emph{in}: the search-based
output types it accepts as inputs.}\label{tab:proto-o}\\
\toprule
\textbf{Prototype} & \textbf{Interaction it captures} & \textbf{out} & \textbf{in} \\
\midrule
\endfirsthead
\toprule
\textbf{Prototype} & \textbf{Interaction it captures} & \textbf{out} & \textbf{in} \\
\midrule
\endhead
\bottomrule
\endfoot
\texttt{flight\_\allowbreak live\_\allowbreak fare} & Live flight fare and itinerary: run a date-, route-, cabin-, and stop-constrained flight search and read the currently displayed itinerary and fare & \texttt{live\_\allowbreak offer} & \texttt{entity},\allowbreak\ \texttt{candidate\_\allowbreak set},\allowbreak\ \texttt{location},\allowbreak\ \texttt{date\_\allowbreak window},\allowbreak\ \texttt{constraint\_\allowbreak set},\allowbreak\ \texttt{identifier} \\
\texttt{stay\_\allowbreak availability} & Live lodging availability: apply destination, dates, party, amenity, and budget constraints and inspect the currently available properties & \texttt{ranked\_\allowbreak item} & \texttt{entity},\allowbreak\ \texttt{candidate\_\allowbreak set},\allowbreak\ \texttt{location},\allowbreak\ \texttt{date\_\allowbreak window},\allowbreak\ \texttt{constraint\_\allowbreak set} \\
\texttt{tour\_\allowbreak slot} & Tour or experience slot: select an experience, date, time, language, and group size and read the available slot and price & \texttt{live\_\allowbreak offer} & \texttt{entity},\allowbreak\ \texttt{candidate\_\allowbreak set},\allowbreak\ \texttt{location},\allowbreak\ \texttt{date\_\allowbreak window},\allowbreak\ \texttt{constraint\_\allowbreak set} \\
\texttt{car\_\allowbreak rental\_\allowbreak quote} & Car-rental quote: enter pickup and return details plus vehicle constraints and read the available rental and full quote & \texttt{quote} & \texttt{entity},\allowbreak\ \texttt{location},\allowbreak\ \texttt{date\_\allowbreak window},\allowbreak\ \texttt{constraint\_\allowbreak set},\allowbreak\ \texttt{identifier} \\
\texttt{event\_\allowbreak ticket} & Event ticket inventory: locate an event and inspect live date, section, quantity, and price inventory & \texttt{live\_\allowbreak offer} & \texttt{entity},\allowbreak\ \texttt{candidate\_\allowbreak set},\allowbreak\ \texttt{location},\allowbreak\ \texttt{date\_\allowbreak window},\allowbreak\ \texttt{constraint\_\allowbreak set},\allowbreak\ \texttt{identifier} \\
\texttt{insurance\_\allowbreak quote} & Insurance quote wizard: complete a profile-, asset-, location-, and coverage-specific quote flow and read the generated estimate & \texttt{quote} & \texttt{entity},\allowbreak\ \texttt{constraint\_\allowbreak set},\allowbreak\ \texttt{identifier},\allowbreak\ \texttt{threshold},\allowbreak\ \texttt{classification} \\
\texttt{finance\_\allowbreak calculator} & Finance calculator result: enter amount, term, rate, contribution, and risk assumptions and read the calculated financial result & \texttt{computed\_\allowbreak result} & \texttt{entity},\allowbreak\ \texttt{constraint\_\allowbreak set},\allowbreak\ \texttt{threshold},\allowbreak\ \texttt{classification} \\
\texttt{product\_\allowbreak configurator} & Product configuration: configure a product on its page for a concrete use case and read the resulting model, compatibility, price, and availability & \texttt{configured\_\allowbreak item} & \texttt{entity},\allowbreak\ \texttt{candidate\_\allowbreak set},\allowbreak\ \texttt{constraint\_\allowbreak set},\allowbreak\ \texttt{identifier},\allowbreak\ \texttt{threshold},\allowbreak\ \texttt{classification} \\
\texttt{auto\_\allowbreak configurator} & Vehicle configuration: select model, powertrain, trim, packages, and appearance options and read the resulting configuration and price & \texttt{configured\_\allowbreak item} & \texttt{entity},\allowbreak\ \texttt{candidate\_\allowbreak set},\allowbreak\ \texttt{constraint\_\allowbreak set},\allowbreak\ \texttt{identifier},\allowbreak\ \texttt{threshold} \\
\texttt{marketplace\_\allowbreak filtered\_\allowbreak item} & Marketplace filtered result: apply category, delivery, seller, condition, rating, and budget constraints and inspect the best displayed result & \texttt{ranked\_\allowbreak item} & \texttt{entity},\allowbreak\ \texttt{candidate\_\allowbreak set},\allowbreak\ \texttt{constraint\_\allowbreak set},\allowbreak\ \texttt{identifier},\allowbreak\ \texttt{threshold},\allowbreak\ \texttt{classification} \\
\texttt{auction\_\allowbreak live\_\allowbreak bid} & Auction state: filter auction lots and inspect the current bid, estimate, time remaining, and lot status & \texttt{live\_\allowbreak status} & \texttt{entity},\allowbreak\ \texttt{candidate\_\allowbreak set},\allowbreak\ \texttt{constraint\_\allowbreak set},\allowbreak\ \texttt{identifier},\allowbreak\ \texttt{threshold},\allowbreak\ \texttt{classification} \\
\texttt{map\_\allowbreak route} & Interactive map route: set endpoints, travel mode, and route constraints and read the rendered route result & \texttt{route} & \texttt{entity},\allowbreak\ \texttt{candidate\_\allowbreak set},\allowbreak\ \texttt{location},\allowbreak\ \texttt{constraint\_\allowbreak set},\allowbreak\ \texttt{identifier} \\
\texttt{transit\_\allowbreak itinerary} & Public-transit itinerary: set stops, departure or arrival time, accessibility, and transfer constraints and read the live itinerary & \texttt{itinerary} & \texttt{entity},\allowbreak\ \texttt{candidate\_\allowbreak set},\allowbreak\ \texttt{location},\allowbreak\ \texttt{date\_\allowbreak window},\allowbreak\ \texttt{constraint\_\allowbreak set},\allowbreak\ \texttt{identifier} \\
\texttt{food\_\allowbreak delivery\_\allowbreak cart} & Food delivery selection: filter restaurants or dishes, configure an order, and read availability, substitutions, fees, and total & \texttt{cart\_\allowbreak total} & \texttt{entity},\allowbreak\ \texttt{candidate\_\allowbreak set},\allowbreak\ \texttt{location},\allowbreak\ \texttt{constraint\_\allowbreak set},\allowbreak\ \texttt{threshold},\allowbreak\ \texttt{classification} \\
\texttt{real\_\allowbreak estate\_\allowbreak filter} & Real-estate filtered property: apply location, price, property, amenity, and commute constraints and inspect matching listings & \texttt{ranked\_\allowbreak item} & \texttt{entity},\allowbreak\ \texttt{candidate\_\allowbreak set},\allowbreak\ \texttt{location},\allowbreak\ \texttt{constraint\_\allowbreak set},\allowbreak\ \texttt{threshold},\allowbreak\ \texttt{classification} \\
\texttt{job\_\allowbreak filter} & Job-board filtered opening: apply role, location, seniority, compensation, skill, and work-mode constraints and inspect matching openings & \texttt{ranked\_\allowbreak item} & \texttt{entity},\allowbreak\ \texttt{candidate\_\allowbreak set},\allowbreak\ \texttt{location},\allowbreak\ \texttt{constraint\_\allowbreak set},\allowbreak\ \texttt{identifier},\allowbreak\ \texttt{threshold},\allowbreak\ \texttt{classification} \\
\texttt{course\_\allowbreak filter} & Course or test-prep selection: apply subject, level, schedule, delivery, prerequisite, and price constraints and inspect the matching course & \texttt{ranked\_\allowbreak item} & \texttt{entity},\allowbreak\ \texttt{candidate\_\allowbreak set},\allowbreak\ \texttt{constraint\_\allowbreak set},\allowbreak\ \texttt{identifier},\allowbreak\ \texttt{threshold},\allowbreak\ \texttt{classification} \\
\texttt{sports\_\allowbreak dashboard} & Sports dashboard state: set league, season, position, scoring, and split controls and read the requested ranking or metric & \texttt{metric} & \texttt{entity},\allowbreak\ \texttt{candidate\_\allowbreak set},\allowbreak\ \texttt{date\_\allowbreak window},\allowbreak\ \texttt{constraint\_\allowbreak set},\allowbreak\ \texttt{identifier},\allowbreak\ \texttt{threshold} \\
\texttt{trading\_\allowbreak screener} & Trading screener state: set market, asset, indicator, horizon, and threshold controls and inspect the resulting asset or chart value & \texttt{metric} & \texttt{entity},\allowbreak\ \texttt{candidate\_\allowbreak set},\allowbreak\ \texttt{date\_\allowbreak window},\allowbreak\ \texttt{constraint\_\allowbreak set},\allowbreak\ \texttt{identifier},\allowbreak\ \texttt{threshold},\allowbreak\ \texttt{classification} \\
\texttt{crypto\_\allowbreak protocol\_\allowbreak quote} & Crypto or DeFi protocol quote: select network, assets, amount, route, and risk constraints and inspect the live swap, bridge, or yield result & \texttt{protocol\_\allowbreak quote} & \texttt{entity},\allowbreak\ \texttt{candidate\_\allowbreak set},\allowbreak\ \texttt{constraint\_\allowbreak set},\allowbreak\ \texttt{identifier},\allowbreak\ \texttt{threshold},\allowbreak\ \texttt{classification} \\
\texttt{template\_\allowbreak filter} & Template gallery selection: filter templates by workflow, audience, features, format, and price and inspect the best matching template & \texttt{template} & \texttt{entity},\allowbreak\ \texttt{candidate\_\allowbreak set},\allowbreak\ \texttt{constraint\_\allowbreak set},\allowbreak\ \texttt{identifier},\allowbreak\ \texttt{classification} \\
\texttt{creative\_\allowbreak visual\_\allowbreak selection} & Creative visual selection: apply style, medium, subject, color, and licensing constraints and inspect a visually matching work & \texttt{visual\_\allowbreak item} & \texttt{entity},\allowbreak\ \texttt{candidate\_\allowbreak set},\allowbreak\ \texttt{constraint\_\allowbreak set},\allowbreak\ \texttt{identifier},\allowbreak\ \texttt{classification} \\
\texttt{gov\_\allowbreak record\_\allowbreak lookup} & Government interactive record lookup: submit jurisdiction, entity, record type, and date constraints and read the returned filing or status & \texttt{record} & \texttt{entity},\allowbreak\ \texttt{candidate\_\allowbreak set},\allowbreak\ \texttt{location},\allowbreak\ \texttt{date\_\allowbreak window},\allowbreak\ \texttt{constraint\_\allowbreak set},\allowbreak\ \texttt{identifier} \\
\texttt{science\_\allowbreak interactive\_\allowbreak result} & Scientific interactive tool result: set dataset, variables, region, transformation, and display controls and read the computed or visualized result & \texttt{computed\_\allowbreak result} & \texttt{entity},\allowbreak\ \texttt{candidate\_\allowbreak set},\allowbreak\ \texttt{location},\allowbreak\ \texttt{date\_\allowbreak window},\allowbreak\ \texttt{constraint\_\allowbreak set},\allowbreak\ \texttt{identifier},\allowbreak\ \texttt{threshold},\allowbreak\ \texttt{classification} \\
\texttt{health\_\allowbreak provider\_\allowbreak filter} & Health-provider finder: apply location, specialty, insurance, availability, and patient constraints and inspect matching providers & \texttt{ranked\_\allowbreak item} & \texttt{entity},\allowbreak\ \texttt{candidate\_\allowbreak set},\allowbreak\ \texttt{location},\allowbreak\ \texttt{date\_\allowbreak window},\allowbreak\ \texttt{constraint\_\allowbreak set},\allowbreak\ \texttt{identifier},\allowbreak\ \texttt{classification} \\
\texttt{home\_\allowbreak service\_\allowbreak quote} & Home-service or utility quote: enter property, location, service, usage, and schedule details and read the generated option or quote & \texttt{service\_\allowbreak quote} & \texttt{entity},\allowbreak\ \texttt{candidate\_\allowbreak set},\allowbreak\ \texttt{location},\allowbreak\ \texttt{constraint\_\allowbreak set},\allowbreak\ \texttt{identifier},\allowbreak\ \texttt{threshold},\allowbreak\ \texttt{classification} \\
\texttt{streaming\_\allowbreak catalog} & Streaming catalog state: apply region, service, genre, language, and maturity constraints and inspect currently available titles & \texttt{ranked\_\allowbreak item} & \texttt{entity},\allowbreak\ \texttt{candidate\_\allowbreak set},\allowbreak\ \texttt{location},\allowbreak\ \texttt{constraint\_\allowbreak set},\allowbreak\ \texttt{identifier},\allowbreak\ \texttt{classification} \\
\texttt{game\_\allowbreak interactive\_\allowbreak build} & Interactive game build or state: set character, mode, equipment, map, and performance constraints and read the resulting build or game state & \texttt{build} & \texttt{entity},\allowbreak\ \texttt{candidate\_\allowbreak set},\allowbreak\ \texttt{constraint\_\allowbreak set},\allowbreak\ \texttt{identifier},\allowbreak\ \texttt{threshold},\allowbreak\ \texttt{classification} \\
\texttt{fitness\_\allowbreak plan\_\allowbreak result} & Fitness or wellness plan result: enter profile, goal, schedule, equipment, and limitation constraints and read the generated class or plan & \texttt{plan} & \texttt{entity},\allowbreak\ \texttt{candidate\_\allowbreak set},\allowbreak\ \texttt{constraint\_\allowbreak set},\allowbreak\ \texttt{identifier},\allowbreak\ \texttt{threshold},\allowbreak\ \texttt{classification} \\
\texttt{saas\_\allowbreak data\_\allowbreak artifact} & SaaS data transformation artifact: configure a data source, transformation, grouping, visualization, and export and inspect the generated artifact & \texttt{artifact} & \texttt{entity},\allowbreak\ \texttt{candidate\_\allowbreak set},\allowbreak\ \texttt{constraint\_\allowbreak set},\allowbreak\ \texttt{identifier},\allowbreak\ \texttt{threshold},\allowbreak\ \texttt{classification} \\
\texttt{flight\_\allowbreak aggregator\_\allowbreak search} & Flight-metasearch itinerary and fare: run a route-, date-, passenger-, cabin-, stop-, and baggage-constrained flight search and inspect a displayed itinerary and fare & \texttt{live\_\allowbreak offer} & \texttt{entity},\allowbreak\ \texttt{candidate\_\allowbreak set},\allowbreak\ \texttt{location},\allowbreak\ \texttt{date\_\allowbreak window},\allowbreak\ \texttt{constraint\_\allowbreak set},\allowbreak\ \texttt{identifier} \\
\texttt{general\_\allowbreak car\_\allowbreak rental} & Car-rental availability and quote: enter pickup and return locations, dates, driver profile, and vehicle constraints and inspect the available rental and complete quote & \texttt{quote} & \texttt{entity},\allowbreak\ \texttt{location},\allowbreak\ \texttt{date\_\allowbreak window},\allowbreak\ \texttt{constraint\_\allowbreak set},\allowbreak\ \texttt{identifier},\allowbreak\ \texttt{classification} \\
\texttt{booking\_\allowbreak tour\_\allowbreak slot} & Travel experience availability: filter travel experiences by destination, activity type, date, time, language, accessibility, and group size and inspect a matching live slot & \texttt{live\_\allowbreak offer} & \texttt{entity},\allowbreak\ \texttt{candidate\_\allowbreak set},\allowbreak\ \texttt{location},\allowbreak\ \texttt{date\_\allowbreak window},\allowbreak\ \texttt{constraint\_\allowbreak set},\allowbreak\ \texttt{identifier} \\
\texttt{cruise\_\allowbreak sailing} & Cruise sailing and cabin availability: search a cruise line by region, embarkation port, sailing window, duration, party, and cabin constraints and inspect a matching live sailing & \texttt{live\_\allowbreak offer} & \texttt{entity},\allowbreak\ \texttt{candidate\_\allowbreak set},\allowbreak\ \texttt{location},\allowbreak\ \texttt{date\_\allowbreak window},\allowbreak\ \texttt{constraint\_\allowbreak set},\allowbreak\ \texttt{identifier} \\
\texttt{intercity\_\allowbreak transport} & Intercity train or bus itinerary: search an intercity rail or coach service using exact stations, date, time, passenger, transfer, and accessibility constraints & \texttt{itinerary} & \texttt{entity},\allowbreak\ \texttt{candidate\_\allowbreak set},\allowbreak\ \texttt{location},\allowbreak\ \texttt{date\_\allowbreak window},\allowbreak\ \texttt{constraint\_\allowbreak set},\allowbreak\ \texttt{identifier} \\
\texttt{ride\_\allowbreak fare\_\allowbreak estimate} & Ride-hail or micromobility estimate: enter pickup, destination, timing, party, and vehicle or accessibility constraints and inspect the displayed ride estimate & \texttt{quote} & \texttt{entity},\allowbreak\ \texttt{candidate\_\allowbreak set},\allowbreak\ \texttt{location},\allowbreak\ \texttt{date\_\allowbreak window},\allowbreak\ \texttt{constraint\_\allowbreak set},\allowbreak\ \texttt{identifier},\allowbreak\ \texttt{threshold} \\
\texttt{urban\_\allowbreak food\_\allowbreak delivery} & Address-specific food delivery cart: set a delivery address, filter restaurants and dishes, configure a cart, and read address-specific availability, fees, and total & \texttt{cart\_\allowbreak total} & \texttt{entity},\allowbreak\ \texttt{candidate\_\allowbreak set},\allowbreak\ \texttt{location},\allowbreak\ \texttt{constraint\_\allowbreak set},\allowbreak\ \texttt{threshold},\allowbreak\ \texttt{classification} \\
\texttt{game\_\allowbreak catalog\_\allowbreak filter} & Game or digital-asset catalog selection: filter a game or digital-asset catalog by platform, genre, mechanics, rating, license, compatibility, and budget and inspect matching listings & \texttt{ranked\_\allowbreak item} & \texttt{entity},\allowbreak\ \texttt{candidate\_\allowbreak set},\allowbreak\ \texttt{constraint\_\allowbreak set},\allowbreak\ \texttt{identifier},\allowbreak\ \texttt{threshold},\allowbreak\ \texttt{classification} \\
\texttt{mortgage\_\allowbreak calculator} & Mortgage payment or rate estimate: enter purchase price, down payment, location, loan term, rate assumptions, taxes, and insurance and inspect the generated mortgage estimate & \texttt{computed\_\allowbreak result} & \texttt{entity},\allowbreak\ \texttt{location},\allowbreak\ \texttt{constraint\_\allowbreak set},\allowbreak\ \texttt{identifier},\allowbreak\ \texttt{threshold},\allowbreak\ \texttt{classification} \\
\texttt{credit\_\allowbreak card\_\allowbreak filter} & Credit-card offer filter: filter credit-card offers by credit profile, fee, reward type, spending pattern, and welcome-offer constraints and inspect matching offers & \texttt{ranked\_\allowbreak item} & \texttt{entity},\allowbreak\ \texttt{candidate\_\allowbreak set},\allowbreak\ \texttt{constraint\_\allowbreak set},\allowbreak\ \texttt{identifier},\allowbreak\ \texttt{threshold},\allowbreak\ \texttt{classification} \\
\texttt{visual\_\allowbreak asset\_\allowbreak filter} & Licensed visual-asset selection: search and visually inspect images by subject, composition, palette, orientation, people, and licensing constraints & \texttt{visual\_\allowbreak item} & \texttt{entity},\allowbreak\ \texttt{candidate\_\allowbreak set},\allowbreak\ \texttt{constraint\_\allowbreak set},\allowbreak\ \texttt{identifier},\allowbreak\ \texttt{classification} \\
\texttt{park\_\allowbreak admission} & Attraction admission and add-on inventory: select an attraction, visit date, party, admission duration, and add-ons and inspect live ticket options and totals & \texttt{live\_\allowbreak offer} & \texttt{entity},\allowbreak\ \texttt{candidate\_\allowbreak set},\allowbreak\ \texttt{location},\allowbreak\ \texttt{date\_\allowbreak window},\allowbreak\ \texttt{constraint\_\allowbreak set},\allowbreak\ \texttt{identifier} \\
\texttt{restaurant\_\allowbreak reservation} & Restaurant table availability: filter restaurants by location, cuisine, dietary support, date, party size, and price and inspect a live reservation slot & \texttt{live\_\allowbreak offer} & \texttt{entity},\allowbreak\ \texttt{candidate\_\allowbreak set},\allowbreak\ \texttt{location},\allowbreak\ \texttt{date\_\allowbreak window},\allowbreak\ \texttt{constraint\_\allowbreak set},\allowbreak\ \texttt{identifier},\allowbreak\ \texttt{classification} \\
\texttt{science\_\allowbreak data\_\allowbreak export} & Scientific dataset subset and export: choose a dataset, variable, spatial and temporal subset, aggregation, units, and export format and inspect the resulting data product & \texttt{artifact} & \texttt{entity},\allowbreak\ \texttt{candidate\_\allowbreak set},\allowbreak\ \texttt{location},\allowbreak\ \texttt{date\_\allowbreak window},\allowbreak\ \texttt{constraint\_\allowbreak set},\allowbreak\ \texttt{identifier},\allowbreak\ \texttt{threshold},\allowbreak\ \texttt{classification} \\
\texttt{defi\_\allowbreak governance\_\allowbreak dashboard} & DeFi governance proposal or gauge state: filter governance proposals, gauges, or votes by protocol, status, round, date, and threshold and inspect the current voting state & \texttt{live\_\allowbreak status} & \texttt{entity},\allowbreak\ \texttt{candidate\_\allowbreak set},\allowbreak\ \texttt{date\_\allowbreak window},\allowbreak\ \texttt{constraint\_\allowbreak set},\allowbreak\ \texttt{identifier},\allowbreak\ \texttt{threshold},\allowbreak\ \texttt{classification} \\
\texttt{collectible\_\allowbreak market\_\allowbreak listing} & Collectible marketplace visual listing: filter collectible listings by object type, period, maker, visual condition, authenticity, seller, auction or fixed-price format, and budget & \texttt{visual\_\allowbreak item} & \texttt{entity},\allowbreak\ \texttt{candidate\_\allowbreak set},\allowbreak\ \texttt{constraint\_\allowbreak set},\allowbreak\ \texttt{identifier},\allowbreak\ \texttt{threshold},\allowbreak\ \texttt{classification} \\
\texttt{game\_\allowbreak item\_\allowbreak catalog} & Official game item or mode catalog: filter the official game catalog or current item and mode listings by game mode, item class, role, compatibility, rotation, and cost & \texttt{ranked\_\allowbreak item} & \texttt{entity},\allowbreak\ \texttt{candidate\_\allowbreak set},\allowbreak\ \texttt{constraint\_\allowbreak set},\allowbreak\ \texttt{identifier},\allowbreak\ \texttt{threshold},\allowbreak\ \texttt{classification} \\
\texttt{fantasy\_\allowbreak lineup} & Fantasy lineup or roster builder: set contest, slate, roster, salary, position, and projection constraints and inspect the resulting lineup or eligible player set & \texttt{build} & \texttt{entity},\allowbreak\ \texttt{candidate\_\allowbreak set},\allowbreak\ \texttt{date\_\allowbreak window},\allowbreak\ \texttt{constraint\_\allowbreak set},\allowbreak\ \texttt{identifier},\allowbreak\ \texttt{threshold},\allowbreak\ \texttt{classification} \\
\end{longtable}

\begin{longtable}{@{}>{\raggedright\arraybackslash}p{3.0cm} >{\raggedright\arraybackslash}p{5.0cm} >{\raggedright\arraybackslash}p{2.75cm} >{\raggedright\arraybackslash}p{2.45cm}@{}}
\caption{Search-based task prototypes $P_{\modeS}$ ($36$ total).
\emph{out}: the prototype's output type. A \emph{producer} supplies an
operation-based input; a \emph{consumer} uses an operation-based output;
cross-site bridges serve both roles.}\label{tab:proto-s}\\
\toprule
\textbf{Prototype} & \textbf{Information need it captures} & \textbf{out} & \textbf{role} \\
\midrule
\endfirsthead
\toprule
\textbf{Prototype} & \textbf{Information need it captures} & \textbf{out} & \textbf{role} \\
\midrule
\endhead
\bottomrule
\endfoot
\texttt{organization\_\allowbreak background} & Organization identity and background: establish who operates the relevant site or entity, where it is based, and the key dated background fact needed by the user & \texttt{fact\_\allowbreak set} & consumer \\
\texttt{entity\_\allowbreak by\_\allowbreak constraints} & Entity identification by public constraints: identify the named entity that satisfies the public, stable constraints in the scenario & \texttt{entity} & producer \\
\texttt{candidate\_\allowbreak shortlist} & Public candidate shortlist: build a short candidate set satisfying stable public criteria before live comparison & \texttt{candidate\_\allowbreak set} & producer \\
\texttt{nearest\_\allowbreak relevant\_\allowbreak location} & Resolve a relevant location: identify the location, hub, venue, district, or waypoint that best satisfies the stated geographic constraint & \texttt{location} & producer \\
\texttt{official\_\allowbreak date\_\allowbreak window} & Resolve an official date window: determine the authoritative event, season, deadline, or eligibility dates needed to parameterize the live task & \texttt{date\_\allowbreak window} & producer \\
\texttt{requirement\_\allowbreak derivation} & Derive requirement set: derive the stable policy, technical, accessibility, dietary, or eligibility requirements that the live result must satisfy & \texttt{constraint\_\allowbreak set} & producer \\
\texttt{public\_\allowbreak identifier} & Resolve public identifier: resolve the canonical identifier, ticker, filing number, model code, station code, or standardized name needed for an exact lookup & \texttt{identifier} & producer \\
\texttt{public\_\allowbreak threshold} & Find a decision threshold: find the authoritative benchmark, limit, target, or recommended threshold to apply in the interactive tool & \texttt{threshold} & producer \\
\texttt{public\_\allowbreak classification} & Resolve classification: determine the official category, risk class, genre, role family, or product class needed to choose the correct interface controls & \texttt{classification} & producer \\
\texttt{official\_\allowbreak policy\_\allowbreak check} & Official policy or eligibility check: use the surfaced entity or result to determine the applicable official policy, eligibility rule, restriction, or obligation & \texttt{fact\_\allowbreak set} & consumer \\
\texttt{external\_\allowbreak benchmark} & External benchmark comparison: compare the surfaced price, quote, metric, route, or calculated value with an authoritative public benchmark on the same basis & \texttt{fact\_\allowbreak set} & consumer \\
\texttt{creator\_\allowbreak owner\_\allowbreak lookup} & Creator, owner, or operator lookup: identify the creator, owner, operator, manufacturer, or issuing authority behind the surfaced item & \texttt{fact\_\allowbreak set} & consumer \\
\texttt{reputation\_\allowbreak and\_\allowbreak reviews} & Independent reputation comparison: use independent public sources to compare reputation, reviews, awards, or reliability for the surfaced option or shortlist & \texttt{fact\_\allowbreak set} & consumer \\
\texttt{historical\_\allowbreak context} & Historical context: place the surfaced entity, metric, status, or result in the relevant historical sequence or prior-period context & \texttt{fact\_\allowbreak set} & consumer \\
\texttt{technical\_\allowbreak compatibility} & Technical compatibility verification: verify from authoritative documentation whether the surfaced configuration, artifact, build, or item is compatible with the user's external system or requirement & \texttt{fact\_\allowbreak set} & consumer \\
\texttt{safety\_\allowbreak accessibility} & Safety and accessibility follow-up: determine the relevant safety, accessibility, health, or operational constraints for the surfaced route, venue, service, or plan & \texttt{fact\_\allowbreak set} & consumer \\
\texttt{availability\_\allowbreak alternative} & Public alternative discovery: find a publicly documented alternative when the surfaced live result is unavailable, over budget, incompatible, or otherwise unsuitable & \texttt{fact\_\allowbreak set} & consumer \\
\texttt{method\_\allowbreak and\_\allowbreak unit\_\allowbreak check} & Method, definition, and unit verification: verify the public definition, method, units, coverage, and limitations needed to interpret the surfaced metric or computed result & \texttt{fact\_\allowbreak set} & consumer \\
\texttt{public\_\allowbreak specification} & Official specification lookup: retrieve the authoritative specifications and limits for the surfaced product, asset, record, or output & \texttt{fact\_\allowbreak set} & consumer \\
\texttt{multi\_\allowbreak source\_\allowbreak fact\_\allowbreak join} & Multi-source enrichment join: join the surfaced entities with the same public attributes from multiple sources and return a comparable table & \texttt{fact\_\allowbreak set} & consumer \\
\texttt{deep\_\allowbreak conjunctive\_\allowbreak entity} & Conjunctive entity identification: identify an initially unknown entity that uniquely satisfies at least four independent clues distributed across multiple sources & \texttt{entity} & producer \\
\texttt{deep\_\allowbreak exhaustive\_\allowbreak shortlist} & Exhaustive constrained shortlist: enumerate a candidate universe, intersect several independently sourced constraints, deduplicate aliases, and return the complete qualifying shortlist & \texttt{candidate\_\allowbreak set} & producer \\
\texttt{deep\_\allowbreak location\_\allowbreak optimization} & Cross-source location resolution: resolve the exact location that best satisfies geographic, jurisdictional, schedule, accessibility, and proximity constraints drawn from multiple sources & \texttt{location} & producer \\
\texttt{deep\_\allowbreak temporal\_\allowbreak window} & Temporal provenance and date-window derivation: reconstruct a dated sequence across authoritative records and derive the exact future window in which all eligibility, event, season, or policy constraints overlap & \texttt{date\_\allowbreak window} & producer \\
\texttt{deep\_\allowbreak requirement\_\allowbreak synthesis} & Multi-authority requirement synthesis: combine policy, technical, legal, accessibility, or compatibility requirements from several authorities into one operational constraint set & \texttt{constraint\_\allowbreak set} & producer \\
\texttt{deep\_\allowbreak identifier\_\allowbreak chain} & Canonical identifier relationship chain: follow a multi-hop relationship chain to resolve the canonical code, registry ID, ticker, model number, station code, or normalized name required for an exact lookup & \texttt{identifier} & producer \\
\texttt{deep\_\allowbreak threshold\_\allowbreak aggregation} & Normalized multi-source threshold: collect comparable numeric evidence from multiple sources, normalize units and time bases, calculate the requested benchmark, and return the decision threshold & \texttt{threshold} & producer \\
\texttt{deep\_\allowbreak classification\_\allowbreak reconciliation} & Cross-authority classification reconciliation: reconcile conflicting taxonomies, definitions, editions, or jurisdictions and determine the one classification applicable to the user's scenario & \texttt{classification} & producer \\
\texttt{deep\_\allowbreak result\_\allowbreak table\_\allowbreak join} & Deep result enrichment table: take every entity surfaced by the operation, resolve stable join keys, collect the same public attributes from multiple sources, and produce a comparable qualified table & \texttt{fact\_\allowbreak set} & consumer \\
\texttt{deep\_\allowbreak source\_\allowbreak reconciliation} & Conflicting-source result reconciliation: use the surfaced entity or value to gather conflicting public claims, compare definitions, dates, scope, units, and authority, and resolve which interpretation applies & \texttt{fact\_\allowbreak set} & consumer \\
\texttt{deep\_\allowbreak historical\_\allowbreak provenance} & Historical provenance of surfaced result: trace the surfaced entity, item, status, or metric through a dated origin, ownership, release, renaming, or prior-period chain and explain the verified transitions & \texttt{fact\_\allowbreak set} & consumer \\
\texttt{deep\_\allowbreak evidence\_\allowbreak synthesis} & Scholarly or technical evidence synthesis: formulate a focused question from the surfaced result, locate multiple primary or review sources, and synthesize methods, findings, disagreement, and limitations & \texttt{fact\_\allowbreak set} & consumer \\
\texttt{cross\_\allowbreak site\_\allowbreak constraint\_\allowbreak bridge} & Cross-site result-to-constraint bridge: use the exact result surfaced on one site to look up the policy, compatibility, eligibility, or comparison rule that becomes a concrete constraint for the next site's operation & \texttt{constraint\_\allowbreak set} & consumer /\allowbreak\ producer \\
\texttt{cross\_\allowbreak site\_\allowbreak identifier\_\allowbreak bridge} & Cross-site canonical-identifier bridge: resolve the surfaced result to the canonical name, code, model, listing identity, or normalized join key required for an exact operation on the next site & \texttt{identifier} & consumer /\allowbreak\ producer \\
\texttt{deep\_\allowbreak cross\_\allowbreak site\_\allowbreak constraint\_\allowbreak bridge} & Deep cross-site requirement reconciliation: take the prior site's exact result, reconcile multi-source policy, technical, legal, or compatibility requirements, and produce one operational constraint set for the next site & \texttt{constraint\_\allowbreak set} & consumer /\allowbreak\ producer \\
\texttt{deep\_\allowbreak cross\_\allowbreak site\_\allowbreak candidate\_\allowbreak bridge} & Deep cross-site candidate transfer: expand and verify the exact entity surfaced on one site against several public sources, reconcile aliases and exclusions, and produce a qualified candidate set for the next site's operation & \texttt{candidate\_\allowbreak set} & consumer /\allowbreak\ producer \\
\end{longtable}
}

\end{document}